\documentclass{article}

    \PassOptionsToPackage{numbers, compress}{natbib}
 \usepackage[main,final]{neurips_2025}

\usepackage{amsmath,amsfonts,bm}

\def\eqref#1{equation~\ref{#1}}
\def\1{\bm{1}}

\DeclareMathAlphabet{\mathsfit}{\encodingdefault}{\sfdefault}{m}{sl}
\SetMathAlphabet{\mathsfit}{bold}{\encodingdefault}{\sfdefault}{bx}{n}

\DeclareMathOperator*{\argmin}{arg\,min}

\usepackage{bm} % fixes boldsymbol
\usepackage{amsmath} % Comment out for acmart
\usepackage{amssymb} % Comment out for acmart
\usepackage{amsthm} % Comment out for acmart
\usepackage[boxed]{algorithm}
\usepackage[noend]{algpseudocode}
\usepackage{braket}

\algnewcommand\algorithmicinput{\textbf{Input:}}
\algnewcommand\Input{\item[\algorithmicinput]}
\algnewcommand\algorithmicoutput{\textbf{Output:}}
\algnewcommand\Output{\item[\algorithmicoutput]}
\algnewcommand\algorithmicalgorithm{\textbf{Algorithm:}}
\algnewcommand\Algorithm{\item[\algorithmicalgorithm]}

\newtheorem{theorem}{Theorem}[section]

\newcommand\veczero{\boldsymbol{\mathrm{0}}}
\newcommand\vecone{\boldsymbol{\mathrm{1}}}

\newcommand\vecq{\boldsymbol{\mathrm{q}}}

\newcommand\vecx{\boldsymbol{\mathrm{x}}}

\newcommand\vecz{\boldsymbol{\mathrm{z}}}

\newcommand\matA{\boldsymbol{\mathrm{A}}}

\newcommand\matD{\boldsymbol{\mathrm{D}}}

\newcommand\matI{\boldsymbol{\mathrm{I}}}

\newcommand\matK{\boldsymbol{\mathrm{K}}}
\newcommand\matL{\boldsymbol{\mathrm{L}}}
\newcommand\matM{\boldsymbol{\mathrm{M}}}

\newcommand\matP{\boldsymbol{\mathrm{P}}}
\newcommand\matQ{\boldsymbol{\mathrm{Q}}}
\newcommand\matR{\boldsymbol{\mathrm{R}}}

\newcommand\matU{\boldsymbol{\mathrm{U}}}
\newcommand\matV{\boldsymbol{\mathrm{V}}}
\newcommand\matW{\boldsymbol{\mathrm{W}}}
\newcommand\matX{\boldsymbol{\mathrm{X}}}

\newcommand\matPi{\boldsymbol{\mathrm{\Pi}}}

\newcommand\matUbar{\overline{\boldsymbol{\mathrm{U}}}}

\theoremstyle{remark}

\newcommand{\birkhoffp}[1]{\Omega_{#1}}				% Birkhoff polytope
\newcommand{\onesv}[1]{\ensuremath{\mathbf{1}_{#1}}}		% Vector of ones (n-dim)
\usepackage[utf8]{inputenc} % allow utf-8 input
\usepackage[T1]{fontenc}    % use 8-bit T1 fonts
\usepackage{url}            % simple URL typesetting
\usepackage{booktabs}       % professional-quality tables
\usepackage{amsfonts}       % blackboard math symbols
\usepackage{nicefrac}       % compact symbols for 1/2, etc.
\usepackage{microtype}      % microtypography
\usepackage{xcolor}         % colors
\definecolor{cvprblue}{rgb}{0.21,0.49,0.74}
\definecolor{lightgrey}{RGB}{150,150,150}
\usepackage[role = cvprblue, grid = lightgrey,skipempty]{credits}

\title{
Quantum Doubly Stochastic Transformers
}

\usepackage[pagebackref,breaklinks,colorlinks,allcolors=blue]{hyperref}
\usepackage{bbm}
\usepackage{comment}

\usepackage{url}
\usepackage{placeins}
\usepackage{mathtools}
\usepackage{enumitem}
\usepackage{caption}
\usepackage{multirow}
\usepackage{multicol}
\usepackage{arydshln} % For dashed lines
\newtheorem{constraint}{Constraint}
\title{Quantum Doubly Stochastic Transformers}

\author{%
  \textbf{Jannis Born} \\[1ex]
  \textbf{Filip Skogh} \quad \quad \textbf{Kahn Rhrissorrakrai} \\[1ex]
  \textbf{Filippo Utro} \quad \quad \textbf{Nico Wagner} \quad \quad \textbf{Aleksandros Sobczyk}\thanks{The author contributed to this work while at IBM Research.} \\[4ex]
  \centerline{\textbf{IBM Research}} \\[4ex]
  \centerline{Correspondence to: \texttt{jab@zurich.ibm.com}}
}

\usepackage{subcaption}
\DeclareCaptionLabelFormat{andtable}{#1~#2\ \&\ \tablename~\thetable}

\begin{document}
\maketitle
\begin{abstract}
At the core of the Transformer, the softmax normalizes the attention matrix to be right stochastic.
Previous research has shown that this often de-stabilizes training and that enforcing the attention matrix to be doubly stochastic (through Sinkhorn's algorithm) consistently improves performance across different tasks, domains and Transformer flavors.
However, Sinkhorn's algorithm is iterative, approximative, non-parametric and thus inflexible w.r.t. the obtained doubly stochastic matrix (DSM). 
Recently, it has been proven that DSMs can be obtained with a parametric quantum circuit, yielding a novel quantum inductive bias for DSMs with no known classical analogue. 
Motivated by this, we demonstrate the feasibility of a hybrid classical-quantum doubly stochastic Transformer (QDSFormer) that replaces the softmax in the self-attention layer with a variational quantum circuit.
We study the expressive power of the circuit and find that it yields more diverse DSMs that better preserve information than classical operators.
Across multiple small-scale object recognition tasks, we find that our QDSFormer consistently surpasses both a standard ViT and other doubly stochastic Transformers. 
Beyond the Sinkformer, this comparison includes a novel quantum-inspired doubly stochastic Transformer (based on QR decomposition) that can be of independent interest.  
Our QDSFormer also shows improved training stability and lower performance variation suggesting that it may mitigate the notoriously unstable training of ViTs on small-scale data.
\end{abstract}
% This is promising because it is currently unknown whether a similarly natural classical approach to produce DSMs parametrically exists.
    
\section{Introduction}
\label{sec:intro}
The Transformer~\citep{vaswani2017attention} continues to be a dominant building block in natural language processing~\citep{dubey2024llama}, computer vision~\citep{dosovitskiy2021an,kirillov2023segment} and biology~\citep{abramson2024accurate}.
Quantum computing (QC), instead, is a novel paradigm with the potential to become practically useful in ML~\citep{schuld2022quantum,havlivcek2019supervised, Liu_2021QuantumSpeedUpML, Huang21_powerofdatainQML,abbas2021power} and fuel applications across disciplines~\citep{basu2023towards,di2024quantum}.
Many attempts have been made to build Transformers with quantum gates, either entirely~\citep{khatri2024quixer,nguyen2024qclusformer,guo2024quantum} or only the attention blocks~\citep{kerenidis2024quantum,evans2024learning,unlu2024hybrid}.
However, rather than merely migrating, recent work in quantum ML identified constraints in specific flavors of neural networks (NN) and successfully mitigated those through quantum -- e.g., fourier NNs~\citep{zhao2024quantum}, graph NNs~\citep{thabet2024quantum} or input-convex NNs~\citep{mariella2024quantum}.
Some known limitations of Transformers are due to the softmax in the attention block, e.g., entropy collapse~\citep{zhai2023stabilizing}, rank collapse~\citep{noci2022signal}, token uniformity~\citep{dong2021attention}, eureka moments~\citep{hoffmann2024eureka} and more~\citep{yang2019mixtape,chang2022softmax,shen2023study,chen2023accumulated,yuan2024towards}.
Applying softmax enforces the attention matrices to be right-stochastic (i.e., rows sum to $1$) while its temperature controls the distribution entropy and is often adjusted to stabilize training~\cite{hoffmann2024eureka,noci2022signal}.

Concurrently, it was discovered that Transformer attention naturally converge to doubly stochastic matrices (DSMs) over training, i.e. their rows \textit{and} columns sum to 1~\cite{sander2022sinkformers}. 
Motivated by this, the \textit{Sinkformer}~\cite{sander2022sinkformers} enforces bistochasticity which boosts Transformer performance across different modalities (text, images, point clouds).
Intuitively, doubly stochastic attention has a similar effect to increasing temperature (entropy) -- attention becomes more "democratic", less interactions are missed and all tokens are being attended more equally.
The Sinkformer~\cite{sander2022sinkformers} is a generalization of Transformers that leverages Sinkhorn's algorithm (SA) and has been widely adopted and extended~\citep{geshkovski2024emergence,ye2024otseg,shahbazi2025espformer}.
Among various techniques to obtain DSMs~\citep{sinkhorn1964relationship,rontsis2020optimal,cuturi2013sinkhorn,wang2010learning}, SA is the most obvious choice, however it has some disadvantages:
\setlist[enumerate,1]{
  labelindent=0pt,    % label flush at paragraph start
  leftmargin=*,       % let enumitem compute labelwidth+labelsep
  listparindent=0pt,  % no extra indent for wrapped text
  itemsep=0pt, parsep=0pt, topsep=0pt
}
\begin{enumerate}
\item It is an iterative approximation procedure which reaches a DSM only in the limit. 
It is thus empirical how many iterations a Sinkformer needs and poor initialization can drastically deteriorate performance~\citep{thornton2023rethinking}.
\item It can guarantee to find a DSMs only if the input matrix is non-negative, which is generally not the case within a Transformer (in practice non-negativity is enforced via exponentiation but we show that this hampers expressivity).
\item  Backpropagating through SA often yields ill-conditioned and exploding/vanishing gradients when 
$\varepsilon$ is small. 
In practice, under early stopping, SA is a sublinearly convergent mirror-descent fixed-point solver rather than a simple, well-conditioned layer~\citep{leger2021gradient}.
\item  It is non-parametric.
Thus, in contrast to e.g., a NN layer, it cannot be optimized regarding \textit{which} DSM should be returned.
\end{enumerate}

Given the empirical superiority of the Sinkformer to vanilla Transformers, it is natural to study different techniques to make attention doubly stochastic.
\begin{figure}[!htb]
    \centering
    \includegraphics[width=.6\linewidth]{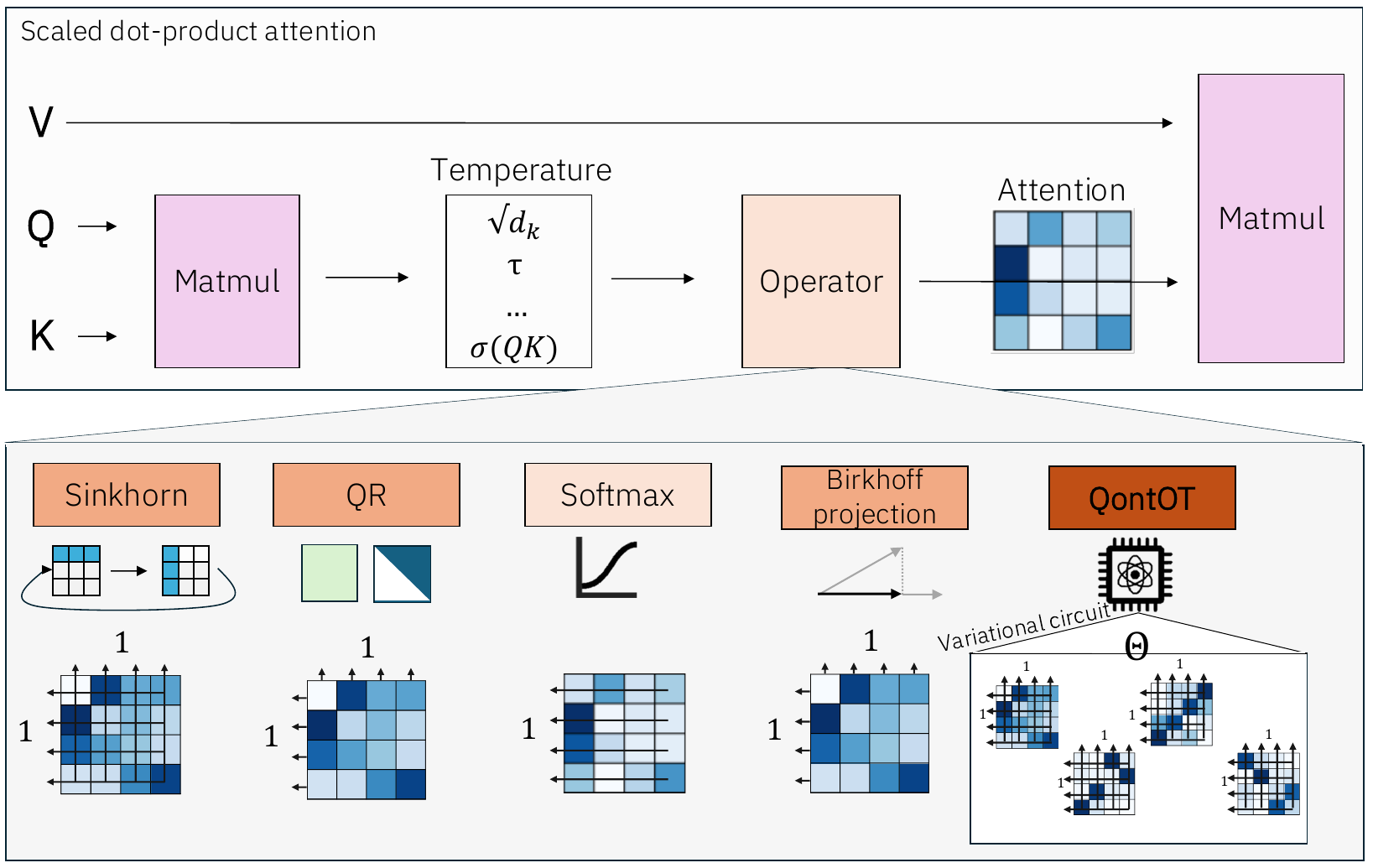}
    \caption{
    \textbf{Doubly Stochastic Transformers.}
    Standard scaled dot-product attention applies a Softmax activation on the query-key matrix (\textit{top}). 
    We study different techniques to make attention
    doubly stochastic attention by replacing the softmax operation (\textit{bottom}).
    Our proposed Quantum Doubly Stochastic Transformer (QDSFormer) leverages QontOT, a variational quantum circuit with high expressivity.
    }
    \label{fig:overview}
    \vspace{-2mm}
\end{figure}
Strikingly, it was recently proven (in a different context) that DSMs can be obtained naturally with a variational (i.e., parametric) quantum circuit, dubbed QontOT~\cite{mariella2024quantum}.
They emphasize that there exists no classical \textit{learning} (i.e., parametric) method that can produce a DSM, akin to QontOT.
Here, we demonstrate that this opens the door for a hybrid quantum-classical doubly-stochastic Transformer (QDSFormer) which offers more flexibility than the Sinkformer.
To that end, we extend QontOT to emit DSMs for an equally-sized matrix.
The resulting quantum layer may replace the softmax within any standard (i.e., non-local, non-sparse) self-attention block.
We focus on replacing the softmax inside the scaled dot-product self-attention of a Vision Transformer (ViT) for three reasons:
\begin{enumerate}[leftmargin=*, topsep=0pt, partopsep=0pt, itemsep=0pt, parsep=0pt]
\item ViTs~\cite{dosovitskiy2021an} suffer from unstable training~\cite{zhai2023stabilizing,hoffmann2024eureka}
\item Unlike in NLP, the attention matrix size is constant, which eases quantum circuit application
\item The attention matrix in a Transformer encoder is unconstrained (unlike in decoders)
\end{enumerate}
% replace the softmax operator with it to compose our QDSFormer.
%%% HERE! mention that we do this in the self-attention block
We empirically analyse expressivity of the quantum circuit, finding that it yields more diverse DSMs than Sinkhorn's algorithm both on synthetic and real data.
It also preserves information better and induces higher entropy.
We then train various flavors of doubly stochastic Transformers (see~\autoref{fig:overview}) on more than ten object recognition datasets.
In comparison to the ViT~\citep{dosovitskiy2021an} and Sinkformer~\cite{sander2022sinkformers}, the QDSFormer shows competitive performance, consistently surpassing both.
In a compositional image recognition task~\cite{hoffmann2024eureka}, we find that they stabilize Transformer training and accelerate learning as they antedate the Eureka moment in compositional problem solving.

In concurrent work, Shahbazi et al. have proposed the EPSFormer~\cite{shahbazi2025espformer} and the LOTFormer~\citep{shahbazi2025lotformer}, two doubly stochastic Transformers that, just like our QDSFormer, overcomes the dependence on Sinkhorn's algorithm to reach doubly stochastic attention.
The ESPFormer~\cite{shahbazi2025espformer} achieves this with sliced OT which is faster than SA but still slower than standard attention. 
Their improvement, the LOTFormer~\citep{shahbazi2025lotformer} marries doubly stochastic and linear attention via conditional OT, yielding better performance and scaling than softmax attention.
Due to their concurrent nature, a performance comparison to ESPFormer and LOTFormer is not included in this work.

\section{Methods}
% \paragraph{Notation}
% We denote the $n$-dimensional vector of ones by $\vecone_n$, the $n\times n$ identity matrix as $\matI_n$ and the $n\times n$ matrix of all ones as $\matJ_n\coloneqq\vecone_n\vecone_n^\top$.
% Given a set $X$, $\convop(X)$ defines its convex hull, i.e., the minimal and unique convex set containing $X$.
% Let $\Delta_{n}$ denote the \textit{probability simplex} in $n - 1$ dimensions, 
% %that is the set
% \begin{align}
% 	\Delta_{n} =& \left\{\vecv \in \mathbb{R}_+^n\middle| \vecone_n^\top\vecv=1\right\}.
% \end{align}

\subsection{Doubly Stochastic Matrices (DSMs)}
We denote the $n$-dimensional vector of ones by $\vecone_n$ and the $n\times n$ identity matrix as $\matI_n$.
The \textit{Birkhoff polytope} $\birkhoffp{n}\coloneqq\mathcal{N}\left(\vecone_n, \vecone_n\right)$~\citep{brualdi_2006} defines the convex set of $n \times n$ doubly stochastic matrices (DSMs). 
A DSM $\matP \in \birkhoffp{n}$ is a non-negative matrix with row/column sum of 1, i.e., 
\begin{align}
    \matP\vecone_n = \vecone_n,\quad&
    \matP^\top\vecone_n = \vecone_n,\quad \matP_{i, j} \ge 0.
\end{align}
A \textit{right stochastic matrix} $\matR$ has \textit{row} sums of 1, i.e., $ \matR\vecone_n = \vecone_n, \quad \matR_{i, j} \ge 0,$
% \begin{align}
%     \matR\vecone_n = \vecone_n, \quad \matR_{i, j} \ge 0,
% \end{align}
and a \textit{left stochastic matrix} $\matL$ has \textit{column} sums of 1, i.e., $\matL^\top\vecone_n = \vecone_n, \quad \matL_{i, j} \ge 0.$
% \begin{align}
%     \matL^\top\vecone_n = \vecone_n, \quad \matL_{i, j} \ge 0.
% \end{align}
Hence, a DSM is left and right stochastic. Moreover, the \textit{Birkhoff-von Neumann theorem} states that the $n!$ vertices (i.e., extreme points) of the Birkhoff polytope $\birkhoffp{n}$ are permutation matrices, so their entries belong to $\{0, 1\}$.
Notably, every DSM $\matP \in \birkhoffp{n}$ can be decomposed as a convex combination of permutation matrices: 
$\matP = \sum_{i=1}^{N} \lambda_i \matPi_i$.
Here $\boldsymbol{\lambda} \in \Delta_N$ is some probability vector in the probability simplex
(denoted as $\Delta_N$), $\{\matPi_i\}$ are the $n\times n$ permutation matrices and $N \le n^2$ denotes the extreme points.
While the decomposition is not unique, each DSM can be represented by at most $n^2$ permutation matrices~\citep{birkhoff1946tres}.
Due to the linear equality constraints, the Birkhoff polytope $\birkhoffp{n}$ lies within a $(n-1)^2$-dimensional affine subspace of the space of $\mathbb{R}^{n\times n}$ matrices.

\subsection{Attention}
We study extensions of dot product attention~\citep{vaswani2017attention}
\begin{equation}
\label{eq:attention}
\text{Attention}(\matQ, \matK, \matV) =  \matA\matV = \text{Softmax}\left(\frac{\matQ\matK^\top}{\tau}\right)\matV 
\end{equation}
where $\matQ\coloneqq \matX\matW_Q$, $\matK\coloneqq \matX\matW_K$ and $\matV\coloneqq \matX\matW_V$ map the input
$\matX$ to query $\matQ$, key $\matK$ and value $\matV$ through their respective weight matrix $\matW_i$ s.t. $\matQ, \matK, \matV\in \mathbb{R}^{T\times d}$.
Moreover, $\tau$ is called the "temperature" and canonically set to $\sqrt{d_k}$~\citep{vaswani2017attention}.
It controls the entropy of the output: low temperature yields a peaky distribution emphasizing differences. 
High temperature attenuates differences thus increasing entropy.
Note that $\tau^{-1} \matQ\matK^\top \in \mathbb{R}^{T \times T}$, so the unnormalized attention matrix is quadratic.
Applying the softmax operator, denoted $ S(\vecz)_i = \frac{\exp(\vecz_i)}{\sum_{j=1}^n \exp(\vecz_j)}$, over the rows makes $\matA$ right-stochastic, i.e., each row $i$ contains a probability distribution denoting the amount of ``attention'' token $i$ pays to the other tokens.
The temperature $\tau$, 

\subsection{Doubly-Stochastic Operators}
\label{sec:operators}
Below we define a non-exhaustive set of operators that can transform $\matM\in \mathbb{R_+}^{T\times T}$ to a DSM $\matP \in \birkhoffp{T}$.
The operators can be integrated into a Transformer by  
$\matM\coloneqq\matQ\matK^\top$ thus yielding a Doubly Stochastic Transformer ("DSFormer").

\subsubsection{Sinkhorn's algorithm}
The most natural approach to obtain a doubly stochastic Transformer was pursued in the Sinkformer~\citep{sander2022sinkformers} and leverages Sinkhorn's algorithm~\citep{sinkhorn1964relationship}. 
Sinkhorn's algorithm is based on Sinkhorn's theorem, stating that for any square strictly positive matrix $\matM\in \mathbb{R_+}^{T\times T}$,
there exist (strictly) positive diagonal matrices $\matP = \matD_1$, $\matD_2$ s.t., $\matD_1\matM\matD_2 \in \birkhoffp{T}$.
Sinkhorn's algorithm, also known as iterative proportional fitting~\citep{bacharach1965estimating}, 
is an approximation procedure that iteratively normalizes the mass of the rows and the columns of $\matM$ which has been proven to converge to a DSM by minimizing Kullback-Leibler (KL) divergence~\citep{soules1991rate}.
The sole hyperparameter of this procedure is $K$, the number of iterations, which we enforce to be odd, following~\cite{sander2022sinkformers},
to ensure the resulting matrix is at least numerically row-stochastic, like for the canonical Softmax operator.
Moreover, we study and compare two implementations of Sinkhorn's algorithm (SA), \texttt{Naive} and \texttt{OT}. \texttt{Naive} alternates between column- and row-normalization: at even iterations ($t$), each column is normalized as $\matP_{ij}^{(t+1)} = \matP_{ij}^{(t)}/{\sum_i \matP_{ij}^{(t)}}$, and at odd iterations, each row is normalized as $\matP_{ij}^{(t+1)} = \matP_{ij}^{(t)}/{\sum_j \matP_{ij}^{(t)}}$.
Instead, the~\texttt{OT} flavor is the operator used in the Sinkformer~\citep{sander2022sinkformers} which relies on the more robust and generalized version to compute optimal transport distances~\cite{cuturi2013sinkhorn}.
Note that, both flavors may not converge with few iterations, especially if $\matQ\matK^\top$ contains large numeric values. 
Therefore, the Sinkformer is only an approximately doubly stochastic Transformer.

\subsubsection{Projection on the Birkhoff polytope}
\label{sec:birkhoff}
Previous work studied different approaches to project matrices onto the Birkhoff polytope~\cite{lim2020doubly,ding2022understanding}, and the most established scheme leverages Frobenius distance~\cite{zass2006doubly}.
Alternatively, one can project $\matM$ directly on $\birkhoffp{T}$ via  $\matP = \underset{\matX \in \birkhoffp{T}}{\argmin} \| \matX-\matM \|_F^2$, where the set for $\matX$ and the objective are convex.
% \TODO{@Alek: Could you please align the notation here with the above? E.g. the vector of ones is defined differently above}
% \begin{align*}
%     \matP = \argmin_{
%         {
%             \substack{
%                 s.t.
%                 \\ 
%                 \matX_{i,j}\geq 0
%                 \\
%                 \matX \vecone_n = \vecone_n
%                 \\
%                 \vecone_n^\top \matX = \vecone_n^\top
%             }
%         }
%     }
%     \| \matX-\matM \|_F^2,
% \end{align*}
We chose to minimize the Frobenius norm here but note that different distances could be explored. 
The resulting problem is a positive-definite convex quadratic program and can be rewritten as
\begin{align}
    \min_{
        {
            \substack{
                s.t.
                \\ 
                \vecx_{i}\geq 0
                \\
                \matA\vecx = \onesv{2n}
            }
        }
    }
    \tfrac{1}{2}\vecx^\top \vecx - \vecq^\top \vecx,
    \quad
    \matA
    =
    {
    \text{\scriptsize
    $
    \begin{pmatrix}
    \vecone_n^\top & \veczero^\top & \ldots   & \veczero^\top \\
    % \veczero_n^\top & \vecone_n^\top & \ldots   & \veczero^\top \\
    \vdots & \vdots  & \ddots   & \vdots \\
    \veczero^\top & \veczero^\top & \ldots   & \vecone_n^\top \\
    \matI_n & \matI_n & \ldots   & \matI_n \\
    \end{pmatrix}
    $}
    }
    % \in\mathbb{R}^{2n\times n^2},
    \label{eq:birkhoff_projection_qp}
\end{align}
%
% \TODO{@Alek: maybe we can condense some of the equations (inline?) to save some space?}
where $\vecx = \mathrm{vec}(\matX^T)$, $\vecq = \mathrm{vec}(\matM^T)$ and $\matA \in \mathbb{R}^{2n\times n^2}$.
The last (or any other) row of $\matA$ can be removed without losing information, since it is a linear combination of the other rows ($\matA$ has rank $2n -1)$~\cite[Thm. 8.1.1]{brualdi_2006}.
We solved the quadratic program with \texttt{OSQP}~\citep{stellato2020osqp}.

\subsubsection{QontOT}
QontOT is a parameterized (variational) quantum circuit that was conceived for conditional prediction of optimal transport plans~\citep{mariella2024quantum} but can be extended to many combinatorial problems~\cite{mermoud2025variational}.
% Importantly, the circuit naturally emits DSMs which are rescaled with classical methods to transport plans.
The circuit naturally emits DSMs and while~\cite{mariella2024quantum} do not find signs of quantum advantage for their main task of optimal transport plan prediction, they report accuracy surpassing their classical baselines for the prediction of DSMs.
This is likely a consequence of the choice of the ansatz which explores a previously unreported link between unitary operators and DSMs.
Indeed, they first proved that DSMs can be obtained naturally with quantum computers thus constructing a quantum inductive bias for DSMs.
Notably, as the authors state, it is currently unknown whether a similarly natural classical approach exists to produce DSMs \textit{parametrically}.

Let $\odot$ be the Hadamard product and $\overline{U}=\left(U^\dagger\right)^\top$ the complex conjugate.
For any unitary matrix $\matU$: $\matU \odot \matUbar \in \birkhoffp{n}$. 
Given the circuit parameters $\theta$ (typically in the hundreds) and $p \in \mathbb{R}$, QontOT obtains a DSM via $\matU(p;\theta) \odot \matUbar(p;\theta)$.
This matrix is block-decomposed before the classical rescaling.
A notable detriment of QontOT is the data injection which is limited to a scalar $p$.
Therefore, we extend the multiplicative data injection $f(\theta, p) = p\cdot\theta$ from scalars to tensors, such that $f(\theta, \matM) = \theta \odot \overrightarrow{\matM}$.
If $\matM$ has less items than $\theta$ we repeat its values to obtain a vector of length identical to $\theta$.
Furthermore, QontOT requires the DSM dimension $n$ to be a power of $2$.
%$\birkhoffp{n}$ where $\log_2{(n)} \in \mathbb{N}$, in other words the DSM dimension must be a power of $2$.
While this may be prohibitive within a Transformer (because sequence length $T$ may differ), it can be mitigated by padding. 
Padding to powers of two is a common technique to maximize hardware efficiency.
Here, we focus our experiments on ViTs because $T$ is a function of patch size.
In general, the circuit size scales favorably in $\mathcal{O}(\log_2(T))$.
It needs at least $4(\log_2(T) + 1)$ qubits, i.e., $2(q_d+q_a+1)$ where $q_d$ is the number of data qubits ($\log_2(T)$) and $q_a$ is the number of auxilliary qubits ($\geq \log_2(T)+1$).
% In practice $T=8$.

\subsubsection{QR Decomposition}
% \TODO{OBC-to-@Everyone: I rewrote a bit this part, commented out the old part. Question: does the procedure remain differentiable under the addition of Gaussian noise?}
% Inspired by the link between unitaries and DSMs~\citep{mariella2024quantum}, we observed that, for a QR decomposition $\matM=\matU\matR$, without loss of generality, $\matU\odot \matU \in \birkhoffp{T}$ is a DSM.
% \footnote{Here we used $\matU$ for the basis matrix of the QR decomposition instead of $\matQ$ to avoid confusion with the query matrix from transformers.} Here $\matU$ is orthogonal and forms an orthonormal basis for the column space of $\matM$, and $\matR$ is upper triangular.
% If the rank is full, this procedure is even differentiable, e.g., via a Gram-Schmidt implementation.
As highlighted above, any unitary $\matU$ can provide a DSM by taking $\matU\odot\matUbar$. 
For any input matrix $\matM$, we can obtain a unitary $\matU$ by computing an orthonormal basis for its column space. 
While there are many ways to obtain a basis, we choose a QR decomposition $\matM=\matU\matR$, in which case $\matR$ is upper triangular.
% \footnote{We use $\matU$ rather than $\matQ$ for the basis of the QR to avoid confusion with the Transformer query matrix.} 
When implemented with Gram-Schmidt, QR is differentiable if $\matM$ is full-rank, but for long-context applications that is rarely the case because query and key matrix have $d = \frac{d_{embed}}{n_{heads}}$ rows and typically $\matM\in\mathbb{R}^{T\times T}$ has rank $\min \{d, T\}$, implying that $\matM$ only has full rank when $d\geq T$.
In practice, if the rank is defective, we inject additive Gaussian noise $\mathcal{N}(0, 1\mathrm{e}{-7})$ to obtain full ranks.
In the ViTs studied in our experiments, $\matM$ often has close to, or full-rank since the dimension $d$ is greater than the number of patches $P$, where $P\approx T$. 
Moreover, QR has time complexity $\mathcal{O}(n^3)$ for dense $n\times n$ matrices, thus to scale up, approximation techniques may be needed~\cite{halko2011finding}.
% @Answer Typically $\matM\in\mathbb{R}^{T\times T}$ has rank $\min \{d, L\}$ where $T$ is the context length. Hence if $d\geq T$, $\matM$ has full rank.} 
% Note that halko2011finding is only faster than Gram-Schmidt if we want to compute a small part of the basis (e.g. a low-rank approximation). If the entire basis is needed, it does not provide an advantage over regular Gram-Schmidt. I am actually not sure if it can be beaten (at least in terms of O(...) notation)

\section{Expressivity of Doubly-Stochastic Operators}
Given the empirical superiority of the Sinkformer to the vanilla (i.e., right-stochastic) Transformer, a natural question is which operator to chose to obtain DSMs.
Before training the DSFormers, we compare the expressivity of the operators -- especially QontOT and Sinkhorn's algorithm -- \textit{in isolation} on synthetic data.
We focus on two aspects. 
\setlist[enumerate,1]{
  labelindent=0pt,    % label flush at paragraph start
  leftmargin=*,       % let enumitem compute labelwidth+labelsep
  listparindent=0pt,  % no extra indent for wrapped text
  itemsep=0pt, parsep=0pt, topsep=0pt
}
\begin{enumerate}
    \item \textbf{Soundness} -- does the operator always produce a DSM?
    Given that $\matU \odot \matUbar \in \birkhoffp{n}$, QontOT always yields a DSM.  
    Similarly for the QR decomposition.
    Instead, Sinkhorn's algorithm (SA) may fail to produce a DSM if the input matrix is not positive. 
    Within the Transformer where $\matM\coloneqq \matQ\matK^\top$, the positivity requirement is generally not fulfilled which is mitigated by input exponentiation.
    Thus, following Sinkhorn's theorem, SA always converges \textit{given} enough iterations $k$. 
    % \TODO{
    % @Nicola: Find references for cases where Sinkhorn does not converge. The latter is more complicated, Nicola will explain this precisely.
    % }
    But in practice the iterative procedure is limiting. 
    When passing $8 \times 8$  $\matQ\matK^\top$ matrices from a trained Sinkformer, we observe that SA does \textit{not} converge for the common choices of $k$ ($3$ and $21$).
    Indeed, the Frobenius distance to the Birkhoff polytope is $0.84_{\pm0.3}$ for $k=3$ and $0.23_{\pm 0.2}$ for $k=21$. 
    This is in contrast to the QR, QontOT and the Birkhoff projection which all yield distances $< 2\mathrm{e}{-4}$, QontOT even $< 5\mathrm{e}{-6}$.
    Instead, the vanilla Softmax operator %(with $\tau=\sqrt{d}$) 
    yields a right-stochastic matrix with distance $1.12_{\pm 0.3}$.
    Hence Sinkformer attention is only approximately doubly stochastic.
    \item \textbf{Completeness} -- can the operator produce all possible DSMs?
    Sinkhorn's algorithm reaches all DSMs of the form $\matP = \matD_1\matM\matD_2$.
    However, due to the entry-wise exponentiation of $\matM$ in the Sinkformer, the input matrix never contains any zero, thus the boundaries of the polytope cannot be reached.
    %\TODO{(Nicola: the latter formula shows that Sinkhorn can reach any DSM, indeed take a DSM and
    %conjugate transform it using two positive diagonal matrices, use the latter as initial matrix for Sinkhorn\ldots)}.
    % \TODO{@Nicola: I dont know whether every point on the polytope can be expressed in this diagonalized form. I searched quickly whether Sinkhorn's algorithm can reach any point on the polytope, but I did not find a conclusive answer. If you know (or could read/think about it), it would help a lot! I thought the extreme points cannot be reached, but are DSMs not fixed points of Sinkhorn's algorithm?}
    Regarding QontOT, the resulting DSM is a convex combination of unistochastic matrices~\cite[Eq. 11b]{mariella2024quantum}.
    Unistochastic matrices are a non-convex proper subset, covering a large amount of the Birkhoff polytope (albeit the exact amount is unknown~\cite{dunkl2009volume}).
    In theory, if all unistochastic matrices could be reached, then by their convex combinations QontOT could cover the entire Birkhoff polyope.
    In practice, reaching all unistochastic matrices (especially all permutation matrices) with the same circuit parametrization is unfeasible as it requires fault-tolerant quantum hardware and high circuit depth (entanglement).
    %) all permutation matrices could be reached and thus the entire Birkhoff polytope would be covered \TODO{Why/how?}
    But over the parameter space of QontOT, the Birkhoff polytope can be approximated more closely.
\end{enumerate}
% First, \textbf{soundness}: does the operator always produce a DSM? Secondly, \textbf{completeness}: Can the operator produce all possible DSMs?
%
% \paragraph{Soundness.}

% $A' = [0 A^T; A 0]$ always converges
%
% \paragraph{Completeness.}

\subsection{Empirical analysis}
\label{sec:expressivity_empirical}
\begin{figure}[!htb]
  \centering
  \begin{subfigure}[b]{0.45\linewidth}
    \centering
    \includegraphics[width=\linewidth]{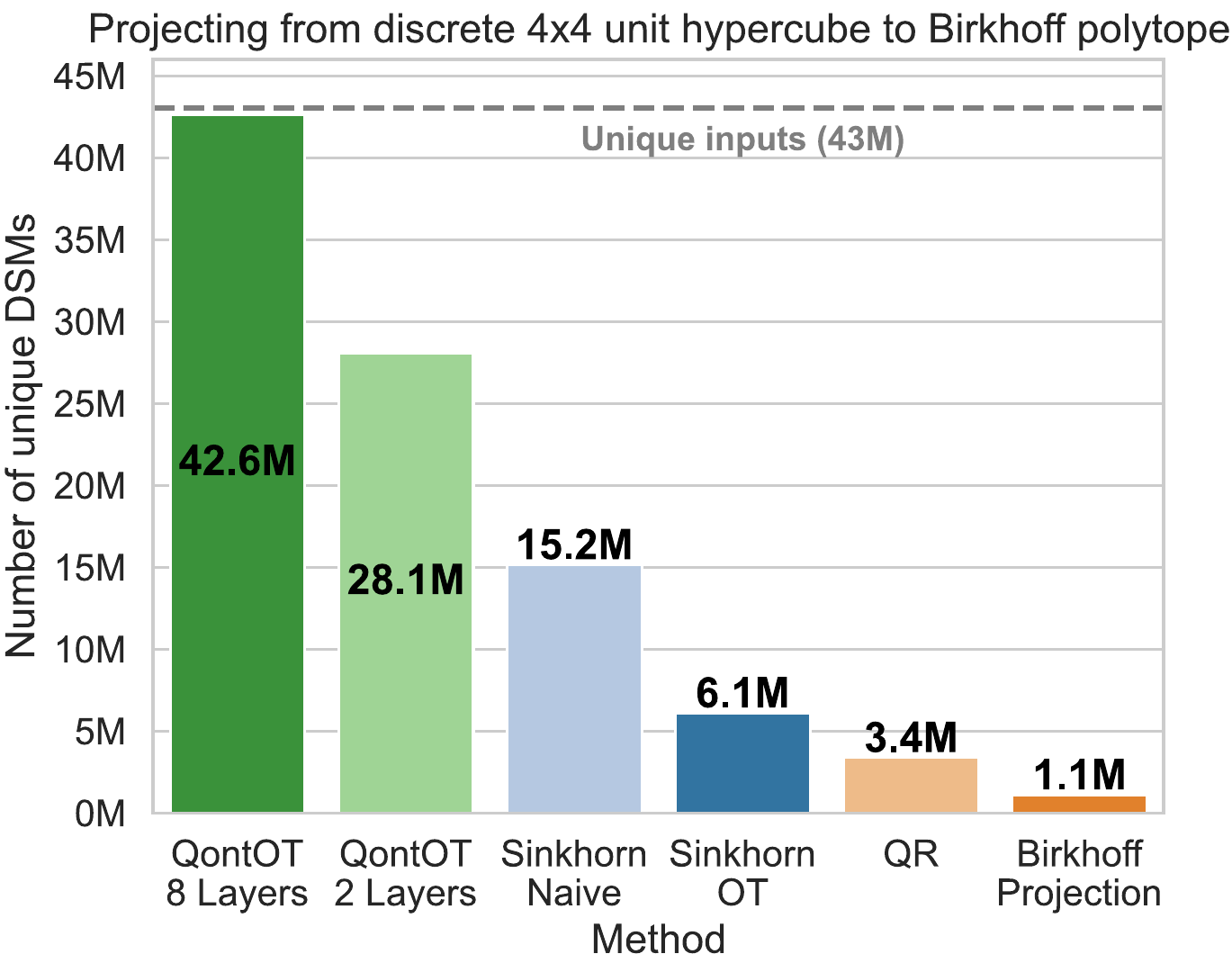}
    \label{fig:expressivity4x4}
  \end{subfigure}
  \hfill
  \begin{subfigure}[b]{0.47\linewidth}
    \centering
    \includegraphics[width=\linewidth]{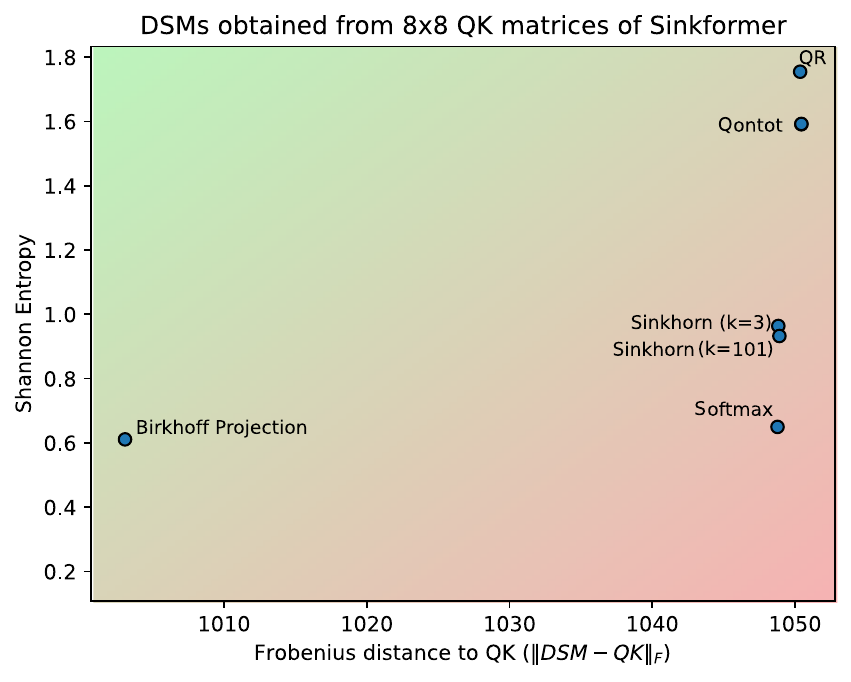}
    \label{fig:8x8expr}
  \end{subfigure}
  \vspace{-5mm}
  \caption{
  \textbf{Left}: Number of unique DSMs obtained after exhaustively iterating over a discretized unit hypercube. With only $8$ layers, QontOT produces a unique DSM for every possible input, unlike all other methods.
  \textbf{Right}: Entropy vs.\ distance-preservation tradeoff. 
  Shannon entropy of different doubly stochastic attention against the Frobenius norm of the difference between unnormalized attention $\matQ\matK$ and the obtained DSM $\matP$. 
  }
  \label{fig:expr}
\end{figure}

% \begin{figure}[!htb]
%     \centering
%     \includegraphics[width=0.6\linewidth]{figures/unique_dsms.pdf}
%     \caption{
%     Number of unique DSMs obtained with each operator after exhaustively iterating over a discretized unit hypercube. 
%     With only $8$ layers, QontOT produces a unique DSM for every possible input, unlike all other methods. 
%     }
%     \label{fig:expressivity4x4}
%     % \vspace{-2mm}
% \end{figure}
To empirically assess the completeness of the operators w.r.t. the Birkhoff polytope, we performed a brute-force analysis over a discretized grid of the unit hypercube. 
For a $n \times n$ matrix and a discretiztion step $d\in \mathbb{N}_+$, we sample each column from a discretized $n$-dimensional hybercube with $d^n$ points, yielding $d^{n^2}$ unique matrices.
% \TODO{@Jannis: The hypercube matrices actually don't have unit length vectors. Suggestion: ``We limit the analysis to matrices with entries in $[0,1]$ because...''}
We refrain from analysing vectors above unit length because all operators are scale-invariant, i.e., $f(\lambda A) = f(A)$.
For $n=4$ and $d=3$ we obtain $3^{16} \approx 43$M matrices 
%(for details see~\autoref{appendix:bruteforce})
and computed the DSM for each input, before rounding to third decimal place.
%Since the studied Birkhoff polytope lives in a $9$-dimensional subspace of $\mathbb{R}^{4\times4}$, it is expected that much less than $43$M unique DSMs are generated.
% The above is wrong because with rounding of 3 decimal places there are still 1000^9 = 10^27 DSMs
Across all operators, QontOT yielded by far the most unique DSMs (see~\autoref{fig:expr}\textbf{A}), behaving nearly injective when $8$ or more circuit layers are used.
This is important, because, none of the other operators is injective 
%across all parts of the input space, 
thus some information is lost when using it inside a neural network.
% The \texttt{OT} implementation of SA and the QR decomposition still yielded $>1$M unique DSMs 
A closer inspection of the empirical cumulative distribution function of all DSMs reveals that QR often emits the same DSMs and that with only $2$ layers (i.e., $98$ parameters), QontOT surpasses all other methods~(see Appendix~\autoref{fig:ecdf}).
Furthermore, whereas all the classical techniques are non-parametric, QontOT yields a different set of DSMs for each parameter configuration.
We repeated above experiment by sampling from a discretized grid of the unit hypersphere (instead of the hypercube).
In this case, Sinkhorn also produces collisions, while QontOT remains injective (there is no proof that the map is injective, but it appears to be experimentaly). 
With a discretization of $d=3$, our sphere contains $625$ matrices where all columns have unit lengths. 
Sinkhorn yields collisions by mapping all rank-1 matrices with constant rows to the center of the Birkhoff polytope, i.e., it fails to differentiate matrix 
$\mathbf{e}_{2}\mathbf{1}^{\top}$ and $\mathbf{e}_{4}\mathbf{1}^{\top}$ ($\mathbf{e}_{2}$ and $\mathbf{e}_{4}$ are the second and forth column of the identity) and thus produces only $621$ unique DSMs whereas QontOT yields $625$ DSMs (QR: 381). This is critical because it implies that Sinkhorn confuses cases where attention matrices are row-wise constant but each row has a unique value. 
%, strengthening its advantage to model the Birkhoff polytope.
%accumulating altogether to the cardinality of the entire Birkhoff polytope.
%%%% THE BELOW IS JUST REMOVED BECAUSE OF PAGE LIMITS
% \begin{comment}
In general, Sinkhorn's algorithm and the direct Birkhoff projection are both permutation and rotation equivariant. 
Instead, QR and QontOT do not possess this characteristic.
Moreover, SA and the QR are scale-invariant 
whereas, again, QontOT is not.
Note that such invariances or equivariances within a Transformer are not generally beneficial or detrimental.
% We also examined the operators on $8\times 8$ $\matQ\matK$ matrices from a single batch ($N=3584$, $7$ heads, batch size of $512$) of a trained Sinkformer.
% These matrices are not generally non-negative and may have large numerical values. 
% In general, the operators were injective on this set, yielding unique DSMs, however the \texttt{OT} flavor of Sinkhorn's algorithm only produced $2175$ ($2021$) unique DSMs for $k=3$ ($21$) iterations.
% Interestingly, the $3584$ unique unnormalized attention matrices from $7$ heads (batch size $512$), the \texttt{OT} flavor of Sinkhorn's Algorithm used within Sinkformer only produced $2175$ ($2021$) unique DSMs for $k=3$ ($21$) iterations.
% Instead, the \texttt{Naive} implementation as well as different flavors of QontOT did not produce any clashes and yielded $3584$ unique DSMs.
% This deficiency may be due to the inputs having large numerical values and not being generally non-negative.

Beyond approximate injectivity, a powerful operator needs to possess two further characteristics: First, information has to be preserved. Obtaining unique DSMs is useless if they destroy information from the input matrix. 
% \begin{figure}[!htb]
%     \centering
%     \includegraphics[width=0.65\linewidth]{figures/attn_sinkhorn_trained_uniform_frob_shannon_entropy.pdf}
%     \caption{
%     Shannon entropy of different flavors of doubly stochastic attention against a measure of distance preservation -- the Frobenius norm of the difference between unnormalized attention $\matQ\matK$ and their obtained DSM $\matP$.
%     QontOT uses $2$ circuit layers.
%     }
%     \label{fig:8x8expr}
%     % \vspace{-2mm}
% \end{figure}
%
%
To assess this, we measured the Frobenius norm of the residuals between input and output matrix\footnote{Other metrics like measuring preservation of ranks or pairwise ratios are possible but yielded similar results.}.
Secondly, low entropy has to be avoided because it causes vanishing gradients and destabilizes Transformer training~\citep{zhai2023stabilizing,shen2023study} for which various mitigation techniques have been suggested~\citep{wang2021escaping,hoffmann2024eureka}.
This so-called ``entropy collapse'' arises if attention is too spiky and is induced by low temperature in the softmax.
Our analysis in~\autoref{fig:expr}\textbf{B} reveals that QontOT possesses comparable information preservation to Sinkhorn while having higher entropy on realistic unnormalized attention matrices.
QR decomposition showed superior entropy, but its cubic scaling limits applicability beyond small-scale Transformers.

Next, we assessed which combination of circuit layers and auxiliary qubits yields the best speed-expressivity compromise. 
In general, a single circuit execution is in the three-digit millisecond range but can be efficiently parallelized.
Increasing the number of layers causes a sub-linear runtime increase, whereas adding more qubits causes a exponential increase (Appendix~\autoref{fig:runtime}). 
Regarding expressivity, adding more layers has a higher impact than adding more qubits (Appendix~\autoref{fig:dsms_range}).
In detail, we passed the same matrix $10,000$ times, sampled the circuit parameters $\theta \sim \mathcal{U}(-1,1)$ and then measured the average range of values covered within each cell of the DSM.
This shows that adding more auxiliary qubits is only useful if even more layers are added simultaneously.
% Altogether, these results are promising and motivate our below empirical experiments with a Quantum Doubly Stochastic Transformer (QDSFormer).
%At the same time, it has to be noted that more entropy is beneficial only within a limited range as close-to-maximal entropy may cause gradients to lower~\citep{noci2022signal}.

\section{Theoretical result on number of DSMs}
\label{sec:dsm}
The optimal way of studying the expressivity of a doubly-stochastic operator empirically would enumerate all DSMs in a given Birkhoff polytope $\Omega_n$ and assess for each DSM whether it can be reached (or how closely).
The exact volume of the Birkhoff polytope is an open problem in mathematics~\cite{chan1999volume} which limits our ability to study expressivity theoretically.
In practice, one can assume a certain discretization $p \in \mathbb{N}_+\text{ s.t., } \matP_{ij} \in \{0, \frac{1}{p-1}, ..., 1\}$, e.g., if $p=2$ then $\matP_{ij} \in \{0,1\}$. 
In that case there are $n!$ DSMs.
%because for the first row there are $n$ columns to place the $1$, for the second row $n-1$ etc.
In~\autoref{sec:dsm_calc} we provide a partial derivation for the combinatorial problem of identifying the function $f(n,p) \rightarrow \mathbb{N}$ returning the number of DSMs.
The basic idea is that a $n \times n$ DSM has $(n-1)^2$ degrees of freedom, thus there are $p^{(n-1)^2}$ candidate matrices.
Not all of these can be turned into DSMs because of two constaints, (1) the sum of any row or column must not exceed 1 and (2) the sum of the $n-1 \times n-1$ submatrix must not be below $n-2$~\cite{chan1999volume}.
This allows to decompose $f$ into $f(n,p) = p^{(n-1)^2} - c_1 - c_2 + c_{12}$ where $c_1$ and $c_2$ measure the constraint violations and $c_{12}$ discounts cases where both constraints are violated.
For details see~\autoref{sec:dsm_calc}.

\section{Quantum Doubly Stochastic Transformer}
\subsection{Experimental Setup}
We evaluate different flavors of DSFormers obtained through replacing the Softmax function with any of the DSM operators described above.
When integrating our QontOT-derived operator into a ViT we obtain our hybrid quantum-classical doubly stochastic Transformer (QDSFormer).
In the following, we refer to QontOT as the attention flavor which contains the quantum circuit whereas QDSFormer denotes, more broadly, any Transformer with quantum doubly-stochastic attention.
To date, the only realization of a QDSFormer is through QontOT.
Among all operators, the classical ones are non-parametric whereas this quantum operator can be optimized during training.
Therefore, one could theoretically optimize circuit parameters concurrently with Transformer training.
However, the ViTs we study contains up to $4$ attention layers, with a batch size of $512$, yielding $2048$ samples to optimize in a single forward pass. 
We predict, unless mentioned otherwise, $8\times 8$ DSMs, use $16$ circuit layers and $4$ auxiliary qubits ($16$ qubits in total).
Running the circuit on quantum hardware requires $\Omega(n^2/\varepsilon^2)$ shots to obtain satisfactory sampling error~\cite{mariella2024quantum}.
Assuming a precision of $\varepsilon=0.01$, this is in the order of $640$k shots per sample.
Since quantum hardware operates on kHz frequency, execution and online optimization on hardware is unfortunately not (yet) feasible.
% As a mitigation strategy we 
%(1) lower precision to $\varepsilon=0.09$, thus we sample only $8$k shots, 
Therefore, we perform exact statevector simulation with Qiskit~\citep{javadi2024quantum} and implement three circuit training strategies:
\begin{enumerate}[leftmargin=*, topsep=0pt, partopsep=0pt, itemsep=0pt, parsep=0pt]
    \item \textbf{Differentiable}: Joint optimization of circuit and Transformer parameters through backpropagation, akin to integrating the circuit as a neural network layer. 
    This is by far the slowest given the difficulty of gradient propagation through quantum circuits~\cite{abbas2024quantum}.
    \item \textbf{Mixed}: A mixed strategy where Transformer training is interleaved with $200$ steps of gradient-free circuit optimization with Nevergrad~\citep{bennet2021nevergrad} on a per-epoch basis.
    \item \textbf{Static}: The circuit is used in pure inference mode with parameters obtained from a 24-qubit DSM prediction experiment on quantum hardware~\cite{mariella2024quantum}.
\end{enumerate}
From the operators studied in~\autoref{sec:expressivity_empirical}, we discard the Birkhoff projection due to its non-differentiability and low DSM-diversity (\autoref{fig:expr}\textbf{A}).
For comparison, we further include the NormSoftmax~\cite{jiang2023normsoftmax}, here denoted as Softmax$_{\sigma}$, that attenuates
attention by taking the minimum of the expected standard deviation $\tau:=\sqrt{d_k}$ and the empirical one:
% \begin{equation}
% \label{eq:normsoftmaxattention}
%\text{Attention} = \text{Softmax}\left(\frac{\matQ\matK^\top}{\min{\left(\sigma(\matQ\matK^\top),\tau\right)}}\right)\matV .
$
\matA = \text{Softmax}\left(\frac{\matQ\matK^\top}{\min{\left(\sigma(\matQ\matK^\top),\tau\right)}}\right)
$.
% \end{equation}
This was found to stabilize ViT training~\cite{jiang2023normsoftmax,hoffmann2024eureka}. Moreover, replacing the standard deviation with the empirical variance, denoted as Softmax$_{\sigma^2}$, improved the performance and stabilized training even more.
Note that both Softmax$_{\sigma}$ and Softmax$_{\sigma^2}$ yields a right-stochastic but not a doubly-stochastic attention matrix.
% we compare NormSoftmax etc, discuss Eureka here
We did not perform hyperparameter optimization for any experiment (for details see Appendix~\ref{sec:hyper}).
We adapted the Sinkformer's ViT implementation of and simply reduced the number of layers and attention heads~\cite{sander2022sinkformers}.

\subsection{Data sets}
We evaluate all ViTs on MNIST~\cite{lecun2010mnist}, Fashion MNIST~\cite{xiao2017fashionmnistnovelimagedataset}, seven datasets from the MedMNIST benchmark~\cite{medmnistv1} and a compositional task requiring multistep reasoning~\cite{hoffmann2024eureka}.
In that task, a $2\times 2$ grid contains two MNIST digits (upper left and lower right) and two FashionMNIST items (upper right and lower left).
If the digits have equal value, the label is the upper right fashion item, otherwise it is the bottom left fashion item.
Performance typically ramps up quickly to $\sim 50\%$ because the model learns to attend one (and \textit{only one}) of the FashionMNIST images. 
Upon continued training with a long saturation phase, a ViT suddenly grasps the relationship of the MNIST digits to the classification task and then climbs rapidly to a $90-95\%$ accuracy.
The moment of abrupt improvement is called "Eureka moment"~\cite{hoffmann2024eureka}.
The dataset is split into $60$K ($10$K) training (validation) examples. 
To accommodate the $8\times 8$ attention matrix, each image from MNIST, FashionMNIST and MedMNIST is split into $7$ horizontal stripes and a \texttt{CLS} token is pre-pended to the patch sequence. 
For the Eureka dataset we use a patch size of $14\times 28$ pixels and mean-pooling.

\subsection{Empirical results}
\begin{figure}[!htb]
    \centering
    \begin{subfigure}[t]{0.35\linewidth}
        \centering
    \includegraphics[width=\linewidth]{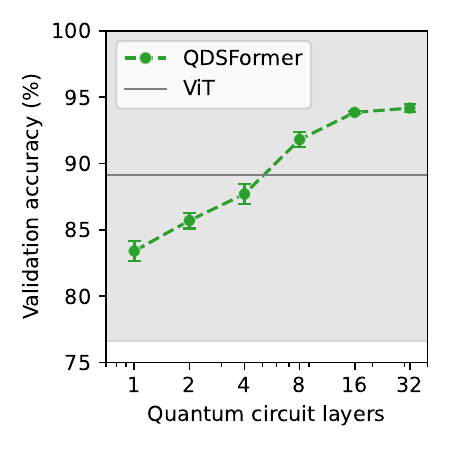}
    \vspace{-6mm}
        \caption{1-ViT-Layer on MNIST.
        }
        \label{fig:1vit-mnist}
    \end{subfigure}
    \hspace{-5mm}
    \hfill
    \begin{subfigure}[t]{0.35\linewidth}
        \centering
        \includegraphics[width=\linewidth]{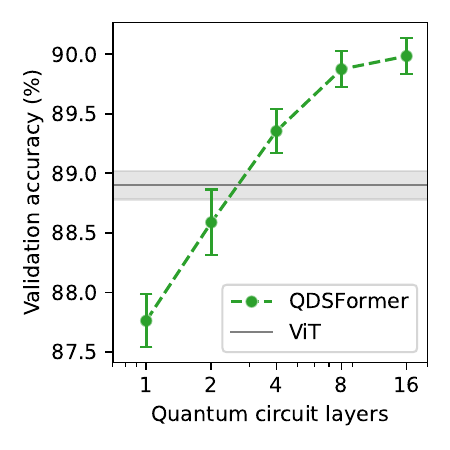}
        \vspace{-6mm}
        \caption{2-ViT-Layer on FashionMNIST.
        }
        \label{fig:2vit-fashionmnist}
    \end{subfigure}
    \vspace{-2mm}
    \caption{
    Comparison of ViT and QDSFormer while varying the circuit depth.
    Mean/std from 5 trainings are shown.
    Within (\textbf{a}) and (\textbf{b}) all models use the same number of trainable parameters.
    }
\label{fig:qdsmformer_layer_ablation}
    \vspace{-2mm}
\end{figure}

First, we compare the QDSFormer directly with a standard ViT.
The ViT uses softmaxed attention whereas the QDSFormer employs a ViT with a static (i.e., non-trainable) instantiation of QontOT to make attention doubly stochastic.
\autoref{fig:qdsmformer_layer_ablation} clearly indicates that on both datasets, the QDSFormer exceeded the ViT by a significant margin. 
This confirms the finding that doubly stochastic attention can improve ViTs~\cite{sander2022sinkformers}. 
Moreover, in both cases, adding more circuit layers increases performance logarithmically and with $4$ or $8$ circuit layers the ViT performance is surpassed.
Exact numerical results are provided in Appendix~\autoref{tab:circuit_layer_ablation}.
Next, we varied the number of ViT layers between $1$ and $4$, comparing to softmaxed attention, Softmax$_{\sigma^2}$~\cite{hoffmann2024eureka} and two classical doubly stochastic attention types: Sinkhorn as used in the Sinkformer~\cite{sander2022sinkformers} and a QR decomposition, a quantum-inspired alternative to QontOT.
All flavors used the same number of parameters and training steps.
On FashionMNIST (\autoref{tab:mnistfashionmnist} \textit{left}) the QDSFormer exceeded all other models for $2$, $3$ and $4$ ViT layers with a performance delta larger than the standard deviation.
%
%
% \begin{table}[!htb]
% \centering
% \caption{
% $L$-layered ViT validation accuracy on FashionMNIST for different attention methods. QontOT uses $16$ circuit layers. Mean/std computed from $5$ trainings.
% }
% \tabcolsep=0.14cm
% \small
% \begin{tabular}{r|ccccc}
% \toprule
% \textit{L} & \textbf{Softmax} & \textbf{Softmax$_{\sigma^2}$} & \textbf{QR} & \textbf{QontOT} & \textbf{Sinkhorn} \\
% \midrule
% 1 & $\underline{86.5}_{\pm 0.19}$ & $75.3_{\pm 4.61}$ & $\textbf{87.1}_{\pm 0.26}$ & $85.6_{\pm 0.10}$ & $84.2_{\pm 3.64}$ \\
% 2 & $88.9_{\pm 0.12}$ & $84.6_{\pm 2.11}$ & $\underline{89.3}_{\pm 0.07}$ & $\textbf{90.0}_{\pm 0.15}$ & $89.1_{\pm 0.73}$ \\
% 3 & $\underline{89.4}_{\pm 0.28}$ & $86.3_{\pm 2.69}$ & $\underline{89.4}_{\pm 0.11}$ & $\textbf{90.3}_{\pm 0.13}$ & $\underline{89.4}_{\pm 0.77}$ \\
% 4 & $\underline{89.7}_{\pm 0.29}$ & $87.1_{\pm 1.15}$ & $89.5_{\pm 0.07}$ & $\textbf{90.3}_{\pm 0.14}$ & $89.1_{\pm 1.08}$ \\
% \bottomrule
% \end{tabular}
% \label{tab:fashionmnist}
% \end{table}
%
The same result was obtained on MNIST and, this time, QontOT outperformed softmaxed attention also for one ViT layer (see~\autoref{tab:mnistfashionmnist} \textit{right}).
%, yet QR and Sinkhorn were superior in this setting.
% Interestingly, on both datasets QR decomposition yielded best performance for $1$ ViT layer.
%
%
% \begin{table}[!htb]
%     \centering
%     \caption{
%     Identical to~\autoref{tab:fashionmnist} but for MNIST.
%     }
%     \tabcolsep=0.14cm
%     \small
%     \begin{tabular}{c|ccccc}
%     \toprule
%     \textbf{$L$} & \textbf{Softmax} & \textbf{Softmax$_{\sigma^2}$} & \textbf{QR} & \textbf{QontOT} & \textbf{Sinkhorn} \\
%     \midrule
%     1 & $89.1_{\pm 12.5}$ & $66.7_{\pm 22.5}$ & $\textbf{96.6}_{\pm 0.10}$ & $93.9_{\pm 0.11}$ & $\underline{94.3}_{\pm 1.97}$ \\
%     2 & $98.1_{\pm 0.33}$ & $93.0_{\pm 4.57}$ & $\underline{98.3}_{\pm 0.13}$ & $\textbf{98.4}_{\pm 0.05}$ & $98.2_{\pm 0.27}$ \\
%     3 & $\underline{98.6}_{\pm 0.11}$ & $97.7_{\pm 0.65}$ & $\underline{98.6}_{\pm 0.13}$ & $\textbf{98.7}_{\pm 0.06}$ & $\underline{98.6}_{\pm 0.12}$ \\
%     4 & $\textbf{98.8}_{\pm 0.10}$ & $97.9_{\pm 0.71}$ & $\underline{98.7}_{\pm 0.11}$ & $\textbf{98.8}_{\pm 0.07}$ & $97.9_{\pm 1.57}$ \\
%     \bottomrule
%     \end{tabular}
%     \label{tab:mnist}
% \end{table}
%
\vspace{-2mm}
\begin{table}[!htb]
  \centering
  \caption{Validation accuracy of $L$-layered ViT on FashionMNIST and MNIST for different attention methods. QontOT uses $16$ circuit layers. Mean/std computed from $5$ trainings.}
  \label{tab:mnistfashionmnist}
  \tabcolsep=0.14cm
  \small
  \resizebox{1.0\columnwidth}{!}{
  \begin{tabular}{c|ccccc|ccccc}
    \toprule
    & \multicolumn{5}{c|}{\textbf{FashionMNIST}} & \multicolumn{5}{c}{\textbf{MNIST}} \\
    \midrule
    \textbf{L}
      & \textbf{Softmax}
      & \textbf{Softmax$_{\sigma^2}$}
      & \textbf{QR}
      & \textbf{QontOT}
      & \textbf{Sinkhorn}
      & \textbf{Softmax}
      & \textbf{Softmax$_{\sigma^2}$}
      & \textbf{QR}
      & \textbf{QontOT}
      & \textbf{Sinkhorn} \\
    \midrule
    1 & $\underline{86.5}_{\pm 0.2}$ & $75.3_{\pm 4.6}$  & $\textbf{87.1}_{\pm 0.3}$ & $85.6_{\pm 0.1}$ & $84.2_{\pm 3.6}$
      & $89.1_{\pm 12.5}$           & $66.7_{\pm 22.5}$ & $\textbf{96.6}_{\pm 0.1}$ & $93.9_{\pm 0.1}$ & $\underline{94.3}_{\pm 2.0}$ \\
    2 & $88.9_{\pm 0.1}$             & $84.6_{\pm 2.1}$  & $\underline{89.3}_{\pm 0.1}$ & $\textbf{90.0}_{\pm 0.2}$ & $89.1_{\pm 0.7}$
      & $98.1_{\pm 0.3}$            & $93.0_{\pm 4.6}$  & $\underline{98.3}_{\pm 0.1}$ & $\textbf{98.4}_{\pm 0.1}$ & $98.2_{\pm 0.3}$ \\
    3 & $\underline{89.4}_{\pm 0.3}$ & $86.3_{\pm 2.7}$  & $\underline{89.4}_{\pm 0.1}$ & $\textbf{90.3}_{\pm 0.1}$ & $\underline{89.4}_{\pm 0.8}$
      & $\underline{98.6}_{\pm 0.1}$ & $97.7_{\pm 0.7}$  & $\underline{98.6}_{\pm 0.1}$ & $\textbf{98.7}_{\pm 0.1}$ & $\underline{98.6}_{\pm 0.1}$ \\
    4 & $\underline{89.7}_{\pm 0.3}$ & $87.1_{\pm 1.2}$  & $89.5_{\pm 0.1}$            & $\textbf{90.3}_{\pm 0.1}$ & $89.1_{\pm 1.1}$
      & $\textbf{98.8}_{\pm 0.1}$    & $97.9_{\pm 0.7}$  & $\underline{98.7}_{\pm 0.1}$ & $\textbf{98.8}_{\pm 0.1}$ & $97.9_{\pm 1.6}$ \\
    \bottomrule
  \end{tabular}
  }
\end{table}

In further experiments with more ViT layers performance assimilated and plateaued due to the simplicity of the datasets.
But we saw scant further improvement for more than $16$ circuit layers.
For a barplot visualization of~\autoref{tab:mnistfashionmnist} see Appendix~\autoref{fig:line_fashion}/\ref{fig:line_mnist}.
Notably, QontOT offers great flexibility in the type of ansatz for the quantum circuit~\cite{mariella2024quantum}. 
We observed only minor differences between four different ansatz types, with three of them outperforming the ViT, underlining the generality of the finding (Appendix~\autoref{tab:ansatztype}).
A compelling aspect is that the static version of the QontOT-attention did perform as good or even better than the optimized one (see Appendix~\autoref{fig:qdsformer_diff}). 
We tested an end-to-end optimizable QDSFormer where circuit  and ViT parameters are jointly optimized.
Such end-to-end training is not only slower, but also had lower accuracy than the static configuration, for both MNIST and FashionMNIST and $1$ and $8$ circuit layers (\autoref{fig:qdsformer_diff}).
This may be caused by Barren plateaus~\cite{mcclean2018barren} (i.e., gradients are largely constant along most directions), a widespread phenomenon in variational quantum circuits that slows down learning.
We further experimented with a "mixed" training strategy where the circuit is trained every $n$-th epoch. 
% More than improving the DSMs produced it also has regularization effects on the model.
This did not reveal a clear benefit for more frequent circuit optimization (see Appendix~\autoref{fig:train_interval}), potentially due to higher volatility of the circuit.
% Table~\ref{tab:circuit_configs} shows how more layers is important for the expressivity circumventing losses in information where two inputs collapse into the same DSM.
We therefore use the static, faster configuration in all remaining experiments.
\begin{table}[!htb]
    \centering
    \caption{
    Test accuracy for MedMNIST datasets across $5$ attention types in a 2-layer ViT. 
    }
    \tabcolsep=0.14cm
    % \small
    % \resizebox{1.0\linewidth}{!}
    {
    \begin{tabular}{c|ccccc}
    \toprule
   \textbf{MedMNIST} & \multirow{2}{*}{\textbf{Softmax}} & \multirow{2}{*}{\textbf{Softmax$_{\sigma^2}$}} & 
   \multirow{2}{*}{\textbf{QR}} & 
   \multirow{2}{*}{\textbf{QontOT}} & 
   \multirow{2}{*}{\textbf{Sinkhorn}} \\
    \textbf{dataset} & & & & & \\
    \midrule
    OCT & $\textbf{64.4}_{\pm 1.6}$ & $43.6_{\pm 3.0}$ & $62.5_{\pm 0.9}$ & $61.6_{\pm 0.6}$ & $55.1_{\pm 5.2}$ \\
    Pneumonia & $84.2_{\pm 0.8}$ & $84.7_{\pm 2.0}$ & $84.3_{\pm 0.7}$ & $\textbf{86.1}_{\pm 1.0}$ & $83.0_{\pm 1.5}$ \\
    Tissue & $60.0_{\pm 0.2}$ & $49.4_{\pm 1.2}$ & $59.0_{\pm 0.1}$ & $\textbf{60.6}_{\pm 0.1}$ & $56.9_{\pm 2.0}$ \\
    OrganA & $78.8_{\pm 0.5}$ & $73.6_{\pm 1.7}$ & $78.4_{\pm 0.6}$ & $\textbf{81.2}_{\pm 0.3}$ & $77.0_{\pm 2.5}$ \\
    OrganC & $79.8_{\pm 0.5}$ & $71.7_{\pm 7.3}$ & $79.6_{\pm 0.3}$ & $\textbf{82.7}_{\pm 0.5}$ & $79.7_{\pm 1.0}$ \\
    OrganS & $64.4_{\pm 0.6}$ & $59.3_{\pm 0.9}$ & $62.6_{\pm 0.8}$ & $\textbf{68.1}_{\pm 0.6}$ & $63.5_{\pm 0.9}$ \\
    Breast & $79.6_{\pm 2.0}$ & $78.2_{\pm 2.2}$ & $\textbf{81.3}_{\pm 2.9}$ & $80.0_{\pm 1.1}$ & $80.1_{\pm 0.8}$ \\
    \bottomrule
    \textbf{Mean} & $73.0$ & $65.8$ & $72.5$ & $\textbf{74.3}$ & $70.8$ \\
    % \textbf{Mean-Std} & $0.89$ & $2.61$ & $0.90$ & $\textbf{0.60}$ & $1.99$ \\
    \bottomrule
    \end{tabular}
    }
    \label{tab:medmnist}
\end{table}
%
% \vspace{-5mm}
Next, we repeated the comparison to the four classical attention types on larger datasets (up to $240$k images) from the MedMNIST benchmark~\cite{medmnistv1}.
In $5$ out of $7$ datasets, the QDSFormer obtained significantly better results than all other methods (\autoref{tab:medmnist}), with a mean accuracy increase of $1.3\%$ compared to a standard ViT.
Notably, none of the other attention types can improve upon the standard ViT.
Another important advantage of the QDSFormer is its stabilizing effect.
Among repeated training runs, the performance variation (i.e., test accuracy variance) is consistently lower than for all classical methods (e.g., \autoref{tab:mnistfashionmnist} or~\autoref{tab:medmnist}). 
Notably, with 1 ViT-layer and softmax-based attention some trainings on MNIST failed to converge.
%For more ViT layers, performance variation  was still significant, in particular for the Sinkformer and among all configurations, the QontOT-based attention had consistently the least variance.
%The Sinkformer was trained with 5 Sinkhorn iterations and we did not see performance gains by further increasing this number.
% A head-to-head comparison to our quantum-\textit{inspired} doubly stochastic attention (QR) yielded mixed results -- QontOT performed better on FashionMNIST, QR was superior on MNIST.
% However, due to the qubic time complexity, scaling QR decomposition to larger attention matrices will be challenging.

\begin{figure}[!htb]
  \centering
  %----------------------------------------------------------------------  
  % make a single line of two top-aligned boxes
  \makebox[\textwidth][c]{%
    \makebox[0.57\columnwidth][t]{%
      \subcaptionbox{Results on the Eureka dataset across different models and learning rates. EM@Ep denotes the average epoch of the EM; runs without EM are set to epoch 100.\label{tab:eureka}}{%
        \resizebox{0.57\columnwidth}{!}{%
        \setlength{\tabcolsep}{2pt}
          \begin{tabular}{lc|ccccc}
            \toprule
            \textbf{LR} & \textbf{Metric}
              & \textbf{Softmax}
              & \textbf{Softmax$_{\sigma^2}$}
              & \textbf{Sinkhorn}
              & \textbf{QR}
              & \textbf{QontOT} \\
            \midrule\midrule
            \multirow{3}{*}{$1$e-$3$}
              & Acc.   & $53.4_{\pm 0.1}$  & $48.6_{\pm 1.1}$  & $49.0_{\pm 0.9}$  & $\underline{61.2}_{\pm 11.0}$  & $\textbf{70.0}_{\pm 13.9}$ \\
              & EM@Ep  & --                & --                & --                & $\textbf{72.2}_{\pm 14.2}$   & $\underline{74.8}_{\pm 8.4}$ \\
              & \# EM  & $0/5$             & $0/5$             & $0/5$             & $\textbf{2/5}$               & $\textbf{2/5}$ \\
            \midrule
            \multirow{3}{*}{$7$e-$4$}
              & Acc.   & $53.9_{\pm 0.2}$  & $50.9_{\pm 0.8}$  & $50.9_{\pm 0.7}$  & $\underline{72.9}_{\pm 10.0}$ & $\textbf{82.3}_{\pm 14.0}$ \\
              & EM@Ep  & --                & --                & --                & $\underline{57.7}_{\pm 11.0}$ & $\textbf{43.3}_{\pm 7.0}$ \\
              & \# EM  & $0/5$             & $0/5$             & $0/5$             & $\textbf{4/5}$               & $\textbf{4/5}$ \\
            \midrule
            \multirow{3}{*}{$5$e-$4$}
              & Acc.   & $61.1_{\pm 15.0}$ & $51.0_{\pm 0.5}$  & $51.6_{\pm 0.2}$  & $\underline{66.4}_{\pm 16.0}$ & $\textbf{89.4}_{\pm 0.1}$ \\
              & EM@Ep  & $61.0$            & --                & --                & $\underline{72.6}_{\pm 22.0}$ & $\textbf{43.6}_{\pm 8.3}$ \\
              & \# EM  & $1/5$             & $0/5$             & $0/5$             & $\underline{2/5}$             & $\textbf{5/5}$ \\
            \bottomrule
          \end{tabular}%
        }%
      }%
    }%
    \hfill
    \makebox[0.42\columnwidth][t]{%
      \subcaptionbox{Validation accuracy per epoch, highlighting the Eureka Moment on the compositional dataset. Confidence bounds from 5 runs.\label{fig:line_eureka}}{%
        \includegraphics[width=0.42\columnwidth]{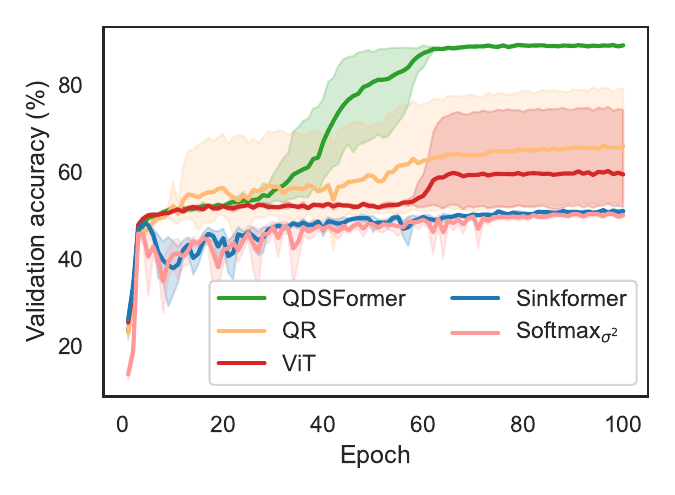}%
      }%
    }%
  }
  %----------------------------------------------------------------------  
  \caption{%
    \textbf{(a)} Eureka results across attention methods.
    \textbf{(b)} QDSFormer antedates the Eureka Moment (EM).
  }
  \label{fig:table_and_eureka_curve}
\end{figure}

Furthermore, to study training stability more systematically, we used a compositional object recognition task with $10$ classes, referred to as "Eureka" dataset.
ViTs are very unstable to train on this task~\cite{hoffmann2024eureka}.
The random seed may determine whether the model saturated at $50\%$ accuracy or experienced a Eureka moment (EM) after hundreds of epochs and would finally converge to $90\%$ accuracy.
As a mitigation strategy, Hoffmann et al.~\cite{hoffmann2024eureka} tame the attention by replacing the Softmax with the NormSoftmax.
In practice, temperature is often tuned manually to find a sweet spot between too low temperature (causing vanishing gradients by low entropy) and too high temperature (causing vanishing gradients by uniform attention).
We speculated that doubly stochastic attention might, \textit{en passant}, antedate the Eureka moment (EM) because it increases attention entropy without making it uniform~\cite{sander2022sinkformers}, thus circumventing temperature tuning.
% \TODO{Mention that in Hoffmann's paper Eureka is often achieved later than 100 epochs. PLus we train on smaller models}
Our experiments confirmed that the standard ViT implementation from~\cite{hoffmann2024eureka} achieves its Eureka moment only after a few hundred epochs.
While the same holds true for their proposed mitigation strategies (Softmax$_{\sigma}$ and Softmax$_{\sigma^2}$), the QDSFormer consistently learned within $100$ epochs to solve this task, resulting in a $30\%$ accuracy improvement over a standard ViT (\autoref{fig:line_eureka}).
%upon removing the exponential moving average (EMA) of the weights, a standard ViT reliably reaches its Eureka moment within the first $10$ epochs thus converging essentially immediately.
% In general we retrain Sinkformer, Norm softmax, and Norm softmax-std from \cite{sander2022sinkformers} to be comparable to our ViT settings.
% This underlines the brittleness of the Eureka phenomenon especially given that, in this scenario, the proposed mitigation techniques (NormSoftmax) perform worst, effectively reversing the trend.
% Earlier Eureka moments were generally accompanied by higher final validation accuracy (see~\autoref{fig:scatter_eureka}).
% The QDSFormer is an exception on that aspect, reaching high accuracy but requiring $30$ epochs to reach the Eureka moment. 
%
%
% \begin{figure}[!htb]
%     \centering
%     \includegraphics[width=0.6\linewidth]{result_figs/eureka_train_line_EM_new.pdf}
%     \vspace{-6mm}
%     \caption{
%     Validation accuracy for each training epoch, highlighting the abrupt learning referred to as Eureka Moment (EM) on the compositional dataset. 
%     Confidence bounds from 5 runs.
%     %, no EMA is used.
%     }
%     \label{fig:line_eureka}
%     % \vspace{-2mm}
% \end{figure}
%
%
This major improvement, achieved with an extremely lightweight quantum circuit (1 layer) also consolidated across different learning rates (\autoref{tab:eureka}).

Next, to assess the potential of the QDSFormer beyond image data and simultaneously study the scalability of the QDSFormer, we applied it on a novel dataset for time-series classification of InfraRed (IR) spectra of molecules into $37$ functional groups~\cite{alberts2025unraveling}.
This dataset contains almost $1$M samples and we scale up the circuit to produce $4$x larger attention matrices ($16\times16$).
On this dataset, the performance differences are marginal and the QDSFormer performs on par with a standard ViT (for details see Appendix~\autoref{tab:spectra}).
This shows that the QDSFormer can meaningfully generalize to domains like scientific data and tasks like multi-label classification.

Since all above results were obtained through statevector simulation, we conducted a final experiment to understand the detrimental effect of quantum noise induced by real quantum hardware via the publicly available \texttt{IBM Quantum Platform}.
We used the three machines \textit{Torino} (Heron R1, 133 qubits, error per layered gate: $1.3\%$), \textit{Brisbane} (Eagle R3, 127 qubits, EPLG: $2.2\%$) and \textit{Cusco} (Eagle R3, 127 qubits, EPLG: $6.8\%$). 
This 14-qubit experiment tests the potential for a hybrid hardware training.
Despite various light error mitigation techniques, the obtained doubly stochastic attention matrices consistently show high entropy (i.e., a tendency toward more uniform distributions), even for larger shot counts.
Experimental details and plots are given in Appendix~\ref{ref:hardware}).
As Appendix~\autoref{fig:hardware}B shows, when comparing to the noise-free ground truth attention matrix, the ordering of the values in the attention matrix was preserved with high precision (spearman $\rho > 0.9$ even for moderate shot count).
Since the circuit runs within a ViT, successfully preserving the ordering will be key (to not destroy signal).
Instead, numerical exactness (cf. Appendix~\autoref{fig:hardware}A) may be compromised: embedding a noisy quantum attention block (which preserves peak attention scores but also increases entropy) into a Transformer could even be advantageous.
The additional entropy may avoid vanishing gradients and act as a form of regularization. 
This effect is particularly notable compared to its noise-free analog, which remains classically intractable if sufficient qubits are used.

\FloatBarrier

\section{Conclusion}
Here, we proposed the Quantum Doubly Stochastic Transformer.
We conceived this method by connecting the centerpiece of a novel variational quantum circuit~\cite{mariella2024quantum} with the Transformer, facilitated through the empirical observation that doubly-stochastic attention improves performance in Transformers~\cite{sander2022sinkformers}.
By extending the QontOT circuit from scalars to matrices we enabled its integration into a ViT, thus providing the first parametric, doubly-stochastic Transformer.
Notably, the QDSFormer presents a meta-class of Sinkformers because it estimates DSMs parametrically, i.e., it can be optimized to learn arbitrary transformations onto the Birkhoff polytope.
Moreover, to our knowledge, there is no classical, parametric approach to estimate DSMs, thus the QDSFormer is a promising candidate for hybrid quantum-classical neural networks trained on quantum hardware.

Our empirical expressivity analysis revealed that the quantum circuit produces DSMs that are more diverse, preserve information better and have higher entropy than DSMs from Sinkhorn's algorithm.
On multiple simple object recognition tasks, the QDSFormer exhibited significantly higher accuracy, outperforming a ViT and a Sinkformer in most cases.
Our usage of quantum attention substantially stabilizes the notoriously unstable ViT training on small-scale data, as evidenced by the performance on a compositional object recognition task (\autoref{fig:line_eureka}),  previously used to study ViT training dynamics~\cite{hoffmann2024eureka}.
Albeit these results are promising, all experiments were performed on comparably small-scale, due to the currently poor scaling of quantum computers in general (which is expected to improve).
Notably, by leveraging QR decomposition, we also proposed a novel, quantum-\textit{inspired} attention flavor.
% On small scale data, this was found to perform extremely well, however scaling it to larger attention matrices will require approximation techniques~\cite{halko2011finding}.
Broadly speaking, outsourcing the activation function to a parametric quantum circuit might be seen a computational overhead, however, we envision that this may reveal potential benefits (typically in small-data, small-model and short-training settings~\cite{abbas2021power}) that are out of reach for classical hardware. 
To that end, future work could explore concurrent optimization of ViT and circuit parameters via the parameter-shift rule on real quantum hardware.
% \begin{equation}
% \label{eq:attention}
% \text{Attention}(\matQ, \matK, \matV) = \mathcal{T}\left(\matQ\matK^\top}\right)\matV 
% \end{equation}
% Typically
% $$\mathcal{T}\left(x\right)=\text{Softmax}\left(\frac{x}{\tau}\right)$$
% where $\matQ, \matK, \matV\in \mathbb{R}^{T\times d}$

\subsection*{Funding Declaration}
This project received no external funding. 

\subsection*{Competing Interests}
The authors declare no competing financial or non-financial interests.

\subsection*{Author Contributions.}
\credit{J.B.}{Software, Conceptualization, Supervision, Formal analysis, Investigation, Writing -- original draft, Visualization, Writing -- review \& editing, Methodology, Data curation, Project administration, Validation}
\credit{F.S}{Data curation, Investigation, Writing -- original draft, Visualization, Writing -- review \& editing, Software, Formal analysis}
\credit{K.R}{Investigation, Visualization, Writing -- review \& editing}
\credit{F.U.}{Investigation, Writing -- review \& editing}
\credit{N.W.}{Investigation, Writing -- original draft}
\credit{A.S.}{Methodology, Formal Analysis, Writing -- original draft, Writing -- review \& editing, Investigation}

\insertcreditsstatement \\
\insertcredits

\FloatBarrier
% \clearpage
% \newpage
{
    \small
    \bibliographystyle{unsrtnat}
    \bibliography{main}

@String(ECCV= {Eur. Conf. Comput. Vis.})

@String(ECCV  = {ECCV})

@inproceedings{ding2022understanding,
  title={Understanding doubly stochastic clustering},
  author={Ding, Tianjiao and Lim, Derek and Vidal, Rene and Haeffele, Benjamin D},
  booktitle={International Conference on Machine Learning},
  pages={5153--5165},
  year={2022},
  organization={PMLR}
}

@article{lim2020doubly,
  title={Doubly stochastic subspace clustering},
  author={Lim, Derek and Vidal, Ren{\'e} and Haeffele, Benjamin D},
  journal={arXiv preprint arXiv:2011.14859},
  year={2020}
}

@article{zass2006doubly,
  title={Doubly stochastic normalization for spectral clustering},
  author={Zass, Ron and Shashua, Amnon},
  journal={Advances in neural information processing systems},
  volume={19},
  year={2006}
}

@article{paszke2017automatic,
  title={Automatic differentiation in pytorch},
  author={Paszke, Adam and Gross, Sam and Chintala, Soumith and Chanan, Gregory and Yang, Edward and DeVito, Zachary and Lin, Zeming and Desmaison, Alban and Antiga, Luca and Lerer, Adam},
  year={2017}
}

@article{mermoud2025variational,
  title={Variational quantum algorithms for permutation-based combinatorial problems: Optimal ansatz generation with applications to quadratic assignment problems and beyond},
  author={Mermoud, Dylan Laplace and Simonetto, Andrea and Elloumi, Sourour},
  journal={arXiv preprint arXiv:2505.05981},
  year={2025}
}

@article{cuturi2013sinkhorn,
  title={Sinkhorn distances: Lightspeed computation of optimal transport},
  author={Cuturi, Marco},
  journal={Advances in neural information processing systems},
  volume={26},
  year={2013}
}

@article{basu2023towards,
  title={Towards quantum-enabled cell-centric therapeutics},
  author={Basu, Saugata and Born, Jannis and Bose, Aritra and Capponi, Sara and Chalkia, Dimitra and Chan, Timothy A and Doga, Hakan and Goldsmith, Mark and Gujarati, Tanvi and Guzman-Saenz, Aldo and others},
  journal={arXiv preprint arXiv:2307.05734},
  year={2023}
}

@InProceedings{mariella2024quantum,
  title = 	 {Quantum Theory and Application of Contextual Optimal Transport},
  author =       {Mariella, Nicola and Akhriev, Albert and Tacchino, Francesco and Zoufal, Christa and Gonzalez-Espitia, Juan Carlos and Harsanyi, Benedek and Koskin, Eugene and Tavernelli, Ivano and Woerner, Stefan and Rapsomaniki, Marianna and Zhuk, Sergiy and Born, Jannis},
  booktitle = 	 {Proceedings of the 41st International Conference on Machine Learning},
  year = 	 {2024},
  volume = 	 {235},
  series = 	 {Proceedings of Machine Learning Research},
  publisher =    {PMLR},
}

@article{sinkhorn1964relationship,
  title={A relationship between arbitrary positive matrices and doubly stochastic matrices},
  author={Sinkhorn, Richard},
  journal={The annals of mathematical statistics},
  volume={35},
  number={2},
  pages={876--879},
  year={1964},
  publisher={JSTOR}
}

@article{Liu_2021QuantumSpeedUpML,
   title={A rigorous and robust quantum speed-up in supervised machine learning},
   ISSN={1745-2481},
   journal={Nature Physics},
   publisher={Springer Science and Business Media LLC},
   author={Liu, Yunchao and Arunachalam, Srinivasan and Temme, Kristan},
   year={2021}
}

@article{di2024quantum,
  title={Quantum Computing for High-Energy Physics: State of the Art and Challenges},
  author={Di Meglio, Alberto and Jansen, Karl and Tavernelli, Ivano and Alexandrou, Constantia and Arunachalam, Srinivasan and Bauer, Christian W and Borras, Kerstin and Carrazza, Stefano and Crippa, Arianna and Croft, Vincent and others},
  journal={PRX Quantum},
  volume={5},
  number={3},
  pages={037001},
  year={2024},
  publisher={APS}
}

@article{abbas2021power,
  title={The power of quantum neural networks},
  author={Abbas, Amira and Sutter, David and Zoufal, Christa and Lucchi, Aur{\'e}lien and Figalli, Alessio and Woerner, Stefan},
  journal={Nature Computational Science},
  volume={1},
  number={6},
  pages={403--409},
  year={2021},
  publisher={Nature Publishing Group US New York}
}

@article{Huang21_powerofdatainQML,
	author = {Huang, Hsin-Yuan and Broughton, Michael and Mohseni, Masoud and Babbush, Ryan and Boixo, Sergio and Neven, Hartmut and McClean, Jarrod R.},
	journal = {Nature Communications},
	number = {1},
	pages = {2631},
	title = {Power of data in quantum machine learning},
	volume = {12},
	year = {2021}}

@article{havlivcek2019supervised,
  title={Supervised learning with quantum-enhanced feature spaces},
  author={Havl{\'\i}{\v{c}}ek, Vojt{\v{e}}ch and C{\'o}rcoles, Antonio D and Temme, Kristan and Harrow, Aram W and Kandala, Abhinav and Chow, Jerry M and Gambetta, Jay M},
  journal={Nature},
  volume={567},
  number={7747},
  pages={209--212},
  year={2019},
  publisher={Nature Publishing Group}
}

@article{schuld2022quantum,
  title={Is quantum advantage the right goal for quantum machine learning?},
  author={Schuld, Maria and Killoran, Nathan},
  journal={Prx Quantum},
  volume={3},
  number={3},
  pages={030101},
  year={2022},
  publisher={APS}
}

@inproceedings{zhai2023stabilizing,
  title={Stabilizing transformer training by preventing attention entropy collapse},
  author={Zhai, Shuangfei and Likhomanenko, Tatiana and Littwin, Etai and Busbridge, Dan and Ramapuram, Jason and Zhang, Yizhe and Gu, Jiatao and Susskind, Joshua M},
  booktitle={International Conference on Machine Learning},
  pages={40770--40803},
  year={2023},
  organization={PMLR}
}

@article{wang2021escaping,
  title={Escaping the gradient vanishing: Periodic alternatives of softmax in attention mechanism},
  author={Wang, Shulun and Liu, Feng and Liu, Bin},
  journal={IEEE Access},
  volume={9},
  pages={168749--168759},
  year={2021},
  publisher={IEEE}
}

@article{noci2022signal,
  title={Signal propagation in transformers: Theoretical perspectives and the role of rank collapse},
  author={Noci, Lorenzo and Anagnostidis, Sotiris and Biggio, Luca and Orvieto, Antonio and Singh, Sidak Pal and Lucchi, Aurelien},
  journal={Advances in Neural Information Processing Systems},
  volume={35},
  pages={27198--27211},
  year={2022}
}

@article{soules1991rate,
  title={The rate of convergence of Sinkhorn balancing},
  author={Soules, George W},
  journal={Linear algebra and its applications},
  volume={150},
  pages={3--40},
  year={1991},
  publisher={Elsevier}
}

@article{geshkovski2024emergence,
  title={The emergence of clusters in self-attention dynamics},
  author={Geshkovski, Borjan and Letrouit, Cyril and Polyanskiy, Yury and Rigollet, Philippe},
  journal={Advances in Neural Information Processing Systems},
  volume={36},
  year={2024}
}

@inproceedings{ye2024otseg,
  title={OTSeg: Multi-prompt Sinkhorn Attention for Zero-Shot Semantic Segmentation},
  author={Ye, Jong Chul and Oh, Yujin and others},
  booktitle={The 18th European Conference on Computer Vision, ECCV 2024},
  year={2024},
  organization={European Computer Vision Association (ECVA)}
}

@article{shen2023study,
  title={A study on relu and softmax in transformer},
  author={Shen, Kai and Guo, Junliang and Tan, Xu and Tang, Siliang and Wang, Rui and Bian, Jiang},
  journal={arXiv preprint arXiv:2302.06461},
  year={2023}
}

@inproceedings{
thabet2024quantum,
title={Quantum Positional Encodings for Graph Neural Networks},
author={Slimane Thabet and Mehdi Djellabi and Igor Olegovich Sokolov and Sachin Kasture and Louis-Paul Henry and Loic Henriet},
booktitle={Forty-first International Conference on Machine Learning},
year={2024},
}

@article{khatri2024quixer,
  title={Quixer: A Quantum Transformer Model},
  author={Khatri, Nikhil and Matos, Gabriel and Coopmans, Luuk and Clark, Stephen},
  journal={arXiv preprint arXiv:2406.04305},
  year={2024}
}

@inproceedings{chang2022softmax,
  title={Softmax bottleneck makes language models unable to represent multi-mode word distributions},
  author={Chang, Haw-Shiuan and McCallum, Andrew},
  booktitle={Proceedings of the 60th Annual Meeting of the Association for Computational Linguistics},
  volume={1},
  year={2022}
}

@inproceedings{chen2023accumulated,
  title={Accumulated trivial attention matters in vision transformers on small datasets},
  author={Chen, Xiangyu and Hu, Qinghao and Li, Kaidong and Zhong, Cuncong and Wang, Guanghui},
  booktitle={Proceedings of the IEEE/CVF Winter Conference on Applications of Computer Vision},
  pages={3984--3992},
  year={2023}
}

@InProceedings{dong2021attention,
  title = 	 {Attention is not all you need: pure attention loses rank doubly exponentially with depth},
  author =       {Dong, Yihe and Cordonnier, Jean-Baptiste and Loukas, Andreas},
  booktitle = 	 {Proceedings of the 38th International Conference on Machine Learning},
  pages = 	 {2793--2803},
  year = 	 {2021},
  editor = 	 {Meila, Marina and Zhang, Tong},
  volume = 	 {139},
  series = 	 {Proceedings of Machine Learning Research},
  month = 	 {18--24 Jul},
  publisher =    {PMLR}
}

@article{yang2019mixtape,
  title={Mixtape: Breaking the softmax bottleneck efficiently},
  author={Yang, Zhilin and Luong, Thang and Salakhutdinov, Russ R and Le, Quoc V},
  journal={Advances in Neural Information Processing Systems},
  volume={32},
  year={2019}
}

@article{nguyen2024qclusformer,
  title={QClusformer: A Quantum Transformer-based Framework for Unsupervised Visual Clustering},
  author={Nguyen, Xuan-Bac and Nguyen, Hoang-Quan and Chen, Samuel Yen-Chi and Khan, Samee U and Churchill, Hugh and Luu, Khoa},
  journal={arXiv preprint arXiv:2405.19722},
  year={2024}
}

@article{unlu2024hybrid,
  title={Hybrid Quantum Vision Transformers for Event Classification in High Energy Physics},
  author={Unlu, Eyup B and Comajoan Cara, Mar{\c{c}}al and Dahale, Gopal Ramesh and Dong, Zhongtian and Forestano, Roy T and Gleyzer, Sergei and Justice, Daniel and Kong, Kyoungchul and Magorsch, Tom and Matchev, Konstantin T and others},
  journal={Axioms},
  volume={13},
  number={3},
  pages={187},
  year={2024},
  publisher={MDPI}
}

@article{guo2024quantum,
  title={Quantum linear algebra is all you need for Transformer architectures},
  author={Guo, Naixu and Yu, Zhan and Choi, Matthew and Agrawal, Aman and Nakaji, Kouhei and Aspuru-Guzik, Al{\'a}n and Rebentrost, Patrick},
  journal={arXiv preprint arXiv:2402.16714},
  year={2024}
}

@article{kerenidis2024quantum,
  title={Quantum vision transformers},
  author={Kerenidis, Iordanis and Mathur, Natansh and Landman, Jonas and Strahm, Martin and Li, Yun Yvonna and others},
  journal={Quantum},
  volume={8},
  pages={1265},
  year={2024},
  publisher={Verein zur F{\"o}rderung des Open Access Publizierens in den Quantenwissenschaften}
}

@article{evans2024learning,
  title={Learning with SASQuaTCh: a Novel Variational Quantum Transformer Architecture with Kernel-Based Self-Attention},
  author={Evans, Ethan N and Cook, Matthew and Bradshaw, Zachary P and LaBorde, Margarite L},
  journal={arXiv preprint arXiv:2403.14753},
  year={2024}
}

@inproceedings{
zhao2024quantum,
title={Quantum Implicit Neural Representations},
author={Jiaming Zhao and Wenbo Qiao and Peng Zhang and Hui Gao},
booktitle={Forty-first International Conference on Machine Learning},
year={2024},
}

@article{javadi2024quantum,
  title={Quantum computing with Qiskit},
  author={Javadi-Abhari, Ali and Treinish, Matthew and Krsulich, Kevin and Wood, Christopher J and Lishman, Jake and Gacon, Julien and Martiel, Simon and Nation, Paul D and Bishop, Lev S and Cross, Andrew W and others},
  journal={arXiv preprint arXiv:2405.08810},
  year={2024}
}

@article{bennet2021nevergrad,
  title={Nevergrad: black-box optimization platform},
  author={Bennet, Pauline and Doerr, Carola and Moreau, Antoine and Rapin, Jeremy and Teytaud, Fabien and Teytaud, Olivier},
  journal={ACM SIGEVOlution},
  volume={14},
  number={1},
  pages={8--15},
  year={2021},
  publisher={ACM New York, NY, USA}
}

@article{bacharach1965estimating,
  title={Estimating nonnegative matrices from marginal data},
  author={Bacharach, Michael},
  journal={International Economic Review},
  volume={6},
  number={3},
  pages={294--310},
  year={1965},
  publisher={JSTOR}
}

@inproceedings{kirillov2023segment,
  title={Segment anything},
  author={Kirillov, Alexander and Mintun, Eric and Ravi, Nikhila and Mao, Hanzi and Rolland, Chloe and Gustafson, Laura and Xiao, Tete and Whitehead, Spencer and Berg, Alexander C and Lo, Wan-Yen and others},
  booktitle={Proceedings of the IEEE/CVF International Conference on Computer Vision},
  pages={4015--4026},
  year={2023}
}

@article{dubey2024llama,
  title={The llama 3 herd of models},
  author={Dubey, Abhimanyu and Jauhri, Abhinav and Pandey, Abhinav and Kadian, Abhishek and Al-Dahle, Ahmad and Letman, Aiesha and Mathur, Akhil and Schelten, Alan and Yang, Amy and Fan, Angela and others},
  journal={arXiv preprint arXiv:2407.21783},
  year={2024}
}

@article{abramson2024accurate,
  title={Accurate structure prediction of biomolecular interactions with AlphaFold 3},
  author={Abramson, Josh and Adler, Jonas and Dunger, Jack and Evans, Richard and Green, Tim and Pritzel, Alexander and Ronneberger, Olaf and Willmore, Lindsay and Ballard, Andrew J and Bambrick, Joshua and others},
  journal={Nature},
  pages={1--3},
  year={2024},
  publisher={Nature Publishing Group UK London}
}

@inproceedings{
dosovitskiy2021an,
title={An Image is Worth 16x16 Words: Transformers for Image Recognition at Scale},
author={Alexey Dosovitskiy and Lucas Beyer and Alexander Kolesnikov and Dirk Weissenborn and Xiaohua Zhai and Thomas Unterthiner and Mostafa Dehghani and Matthias Minderer and Georg Heigold and Sylvain Gelly and Jakob Uszkoreit and Neil Houlsby},
booktitle={International Conference on Learning Representations},
year={2021},
url={https://openreview.net/forum?id=YicbFdNTTy}
}

@inproceedings{rontsis2020optimal,
  title={Optimal approximation of doubly stochastic matrices},
  author={Rontsis, Nikitas and Goulart, Paul},
  booktitle={International Conference on Artificial Intelligence and Statistics},
  pages={3589--3598},
  year={2020},
  organization={PMLR}
}

@InProceedings{hoffmann2024eureka,
  title = 	 {Eureka-Moments in Transformers: Multi-Step Tasks Reveal Softmax Induced Optimization Problems},
  author =       {Hoffmann, David T and Schrodi, Simon and Bratuli\'{c}, Jelena and Behrmann, Nadine and Fischer, Volker and Brox, Thomas},
  booktitle = 	 {Proceedings of the 41st International Conference on Machine Learning},
  pages = 	 {18409--18438},
  year = 	 {2024},
  volume = 	 {235},
  series = 	 {Proceedings of Machine Learning Research},
  month = 	 {21--27 Jul},
  publisher =    {PMLR}
}

@article{birkhoff1946tres,
  title={Tres observaciones sobre el algebra lineal},
  author={Birkhoff, Garrett},
  journal={Univ. Nac. Tucuman, Ser. A},
  volume={5},
  pages={147--154},
  year={1946}
}

@inproceedings{sander2022sinkformers,
  title={Sinkformers: Transformers with doubly stochastic attention},
  author={Sander, Michael E and Ablin, Pierre and Blondel, Mathieu and Peyr{\'e}, Gabriel},
  booktitle={International Conference on Artificial Intelligence and Statistics},
  pages={3515--3530},
  year={2022},
  organization={PMLR}
}

@book{brualdi_2006,
	place={Cambridge},
	series={Encyclopedia of Mathematics and its Applications},
	title={Combinatorial Matrix Classes},
	publisher={Cambridge University Press},
	author={Brualdi, Richard A.},
	year={2006},
	collection={Encyclopedia of Mathematics and its Applications}
}

@article{vaswani2017attention,
  title={Attention is all you need},
  author={Vaswani, Ashish and Shazeer, Noam and Parmar, Niki and Uszkoreit, Jakob and Jones, Llion and Gomez, Aidan N and others},
  journal={Advances in neural information processing systems},
  volume={30},
  number={1},
  pages={261--272},
  year={2017}
}

@article{stellato2020osqp,
  author  = {Stellato, B. and Banjac, G. and Goulart, P. and Bemporad, A. and Boyd, S.},
  title   = {{OSQP}: an operator splitting solver for quadratic programs},
  journal = {Mathematical Programming Computation},
  volume  = {12},
  number  = {4},
  pages   = {637--672},
  year    = {2020},
  doi     = {10.1007/s12532-020-00179-2},
  url     = {https://doi.org/10.1007/s12532-020-00179-2},
}

@article{mcclean2018barren,
  title={Barren plateaus in quantum neural network training landscapes},
  author={McClean, Jarrod R and Boixo, Sergio and Smelyanskiy, Vadim N and Babbush, Ryan and Neven, Hartmut},
  journal={Nature communications},
  volume={9},
  number={1},
  pages={4812},
  year={2018},
  publisher={Nature Publishing Group UK London}
}

@misc{xiao2017fashionmnistnovelimagedataset,
      title={Fashion-MNIST: a Novel Image Dataset for Benchmarking Machine Learning Algorithms}, 
      author={Han Xiao and Kashif Rasul and Roland Vollgraf},
      year={2017},
      eprint={1708.07747},
      archivePrefix={arXiv},
      primaryClass={cs.LG},
      url={https://arxiv.org/abs/1708.07747}, 
}

@article{lecun2010mnist,
  title={MNIST handwritten digit database},
  author={LeCun, Yann and Cortes, Corinna and Burges, CJ},
  journal={ATT Labs [Online]. Available: http://yann.lecun.com/exdb/mnist},
  volume={2},
  year={2010}
}

@inproceedings{jiang2023normsoftmax,
  title={NormSoftmax: Normalizing the Input of Softmax to Accelerate and Stabilize Training},
  author={Jiang, Zixuan and Gu, Jiaqi and Pan, David Z},
  booktitle={2023 IEEE International Conference on Omni-layer Intelligent Systems (COINS)},
  pages={1--6},
  year={2023},
  organization={IEEE}
}

@article{abbas2024quantum,
  title={On quantum backpropagation, information reuse, and cheating measurement collapse},
  author={Abbas, Amira and King, Robbie and Huang, Hsin-Yuan and Huggins, William J and Movassagh, Ramis and Gilboa, Dar and McClean, Jarrod},
  journal={Advances in Neural Information Processing Systems},
  volume={36},
  year={2024}
}

@inproceedings{medmnistv1,
    title={MedMNIST Classification Decathlon: A Lightweight AutoML Benchmark for Medical Image Analysis},
    author={Yang, Jiancheng and Shi, Rui and Ni, Bingbing},
    booktitle={IEEE 18th International Symposium on Biomedical Imaging (ISBI)},
    pages={191--195},
    year={2021}
}

@article{khatri2019quantum,
  title={Quantum-assisted quantum compiling},
  author={Khatri, Sumeet and LaRose, Ryan and Poremba, Alexander and Cincio, Lukasz and Sornborger, Andrew T and Coles, Patrick J},
  journal={Quantum},
  volume={3},
  pages={140},
  year={2019},
  publisher={Verein zur F{\"o}rderung des Open Access Publizierens in den Quantenwissenschaften}
}

@article{smith2019simulating,
  title={Simulating quantum many-body dynamics on a current digital quantum computer},
  author={Smith, Adam and Kim, MS and Pollmann, Frank and Knolle, Johannes},
  journal={npj Quantum Information},
  volume={5},
  number={1},
  pages={106},
  year={2019},
  publisher={Nature Publishing Group UK London}
}

@article{wallman2016noise,
  title={Noise tailoring for scalable quantum computation via randomized compiling},
  author={Wallman, Joel J and Emerson, Joseph},
  journal={Physical Review A},
  volume={94},
  number={5},
  pages={052325},
  year={2016},
  publisher={APS}
}

@article{ezzell2023dynamical,
  title={Dynamical decoupling for superconducting qubits: a performance survey},
  author={Ezzell, Nic and Pokharel, Bibek and Tewala, Lina and Quiroz, Gregory and Lidar, Daniel A},
  journal={Physical Review Applied},
  volume={20},
  number={6},
  pages={064027},
  year={2023},
  publisher={APS}
}

@article{madden2022best,
  title={Best approximate quantum compiling problems},
  author={Madden, Liam and Simonetto, Andrea},
  journal={ACM Transactions on Quantum Computing},
  volume={3},
  number={2},
  pages={1--29},
  year={2022},
  publisher={ACM New York, NY}
}

@article{chan1999volume,
  title={On the volume of the polytope of doubly stochastic matrices},
  author={Chan, Clara S and Robbins, David P},
  journal={Experimental Mathematics},
  volume={8},
  number={3},
  pages={291--300},
  year={1999},
  publisher={Taylor \& Francis}
}

@article{halko2011finding,
  title={Finding structure with randomness: Probabilistic algorithms for constructing approximate matrix decompositions},
  author={Halko, Nathan and Martinsson, Per-Gunnar and Tropp, Joel A},
  journal={SIAM review},
  volume={53},
  number={2},
  pages={217--288},
  year={2011},
  publisher={SIAM}
}

@article{alberts2025unraveling,
  title={Unraveling Molecular Structure: A Multimodal Spectroscopic Dataset for Chemistry},
  author={Alberts, Marvin and Schilter, Oliver and Zipoli, Federico and Hartrampf, Nina and Laino, Teodoro},
  journal={Advances in Neural Information Processing Systems},
  volume={37},
  pages={125780--125808},
  year={2025}
}

@article{yuan2024towards,
  title={Towards Better Multi-head Attention via Channel-wise Sample Permutation},
  author={Yuan, Shen and Xu, Hongteng},
  journal={arXiv preprint arXiv:2410.10914},
  year={2024}
}

@article{shahbazi2025lotformer,
  title={LOTFormer: Doubly-Stochastic Linear Attention via Low-Rank Optimal Transport},
  author={Shahbazi, Ashkan and Thrash, Chayne and Bai, Yikun and Hamm, Keaton and NaderiAlizadeh, Navid and Kolouri, Soheil},
  journal={arXiv preprint arXiv:2509.23436},
  year={2025}
}

@inproceedings{thornton2023rethinking,
  title={Rethinking initialization of the sinkhorn algorithm},
  author={Thornton, James and Cuturi, Marco},
  booktitle={International Conference on Artificial Intelligence and Statistics},
  pages={8682--8698},
  year={2023},
  organization={PMLR}
}

@article{leger2021gradient,
  title={A gradient descent perspective on Sinkhorn},
  author={L{\'e}ger, Flavien},
  journal={Applied Mathematics \& Optimization},
  volume={84},
  number={2},
  pages={1843--1855},
  year={2021},
  publisher={Springer}
}

@inproceedings{
    shahbazi2025espformer,
    title={{ESPF}ormer: Doubly-Stochastic Attention with Expected Sliced Transport Plans},
    author={Ashkan Shahbazi and Elaheh Akbari and Darian Salehi and Xinran Liu and Navid NaderiAlizadeh and Soheil Kolouri},
    booktitle={Forty-second International Conference on Machine Learning},
    year={2025}
}

@inproceedings{wang2010learning,
  title={Learning a bi-stochastic data similarity matrix},
  author={Wang, Fei and Li, Ping and Konig, Arnd Christian},
  booktitle={2010 IEEE International Conference on Data Mining},
  pages={551--560},
  year={2010},
  organization={IEEE}
}

@article{dunkl2009volume,
  title={Volume of the set of unistochastic matrices of order 3 and the mean Jarlskog invariant},
  author={Dunkl, Charles and {\.Z}yczkowski, Karol},
  journal={Journal of mathematical physics},
  volume={50},
  number={12},
  year={2009},
  publisher={AIP Publishing}
}

@misc{rw2019timm,
  author = {Ross Wightman},
  title = {PyTorch Image Models},
  year = {2019},
  publisher = {GitHub},
  journal = {GitHub repository},
  doi = {10.5281/zenodo.4414861},
  howpublished = {\url{https://github.com/rwightman/pytorch-image-models}}
}

@article{Kingma2014AdamAM,
  title={Adam: A Method for Stochastic Optimization},
  author={Diederik P. Kingma and Jimmy Ba},
  journal={CoRR},
  year={2014},
  volume={abs/1412.6980},
  url={https://api.semanticscholar.org/CorpusID:6628106}
}

@inproceedings{
loshchilov2018adamW,
title={Decoupled Weight Decay Regularization},
author={Ilya Loshchilov and Frank Hutter},
booktitle={International Conference on Learning Representations},
year={2019},
url={https://openreview.net/forum?id=Bkg6RiCqY7},
}
}

 \renewcommand{\thefigure}{A\arabic{figure}}
\setcounter{figure}{0}
 \renewcommand{\thetable}{A\arabic{table}}
\setcounter{table}{0}

\appendix
\clearpage
\newpage

% \section{Generating input matrices for the experiments}
% \label{appendix:bruteforce}
% For the empirical expressivity analysis we brute-forced over all possible matrices in a hypercube.
% %we used two different methods to generate test matrices. 
% % \subsection{Brute forcing all possible matrices over a hypercube}
% We choose a matrix size $n\in\mathbb{N}$ and a discretiztion step $d\in \mathbb{N}_+$. Each of the $n$ columns is sampled from a discretized $n$-dimensional hybercube with $d^n$ points in total. There are $n$ columns, which gives a total of $d^{n^2}$ different matrices in total.

% \subsection{Sampling from the unit sphere}
% To complement the analysis, we also sample matrices uniformly from the unit sphere. Specifically, we sample $k$ vectors $\vecg_i,i\in[k]$, where each vector has $n^2$ i.i.d elements from $\mathcal{N}(0,1)$. The vectors are then scaled to have unit norm, and flattened row-wise to form an input matrix ($k$ matrices in total).

%
%
% \onecolumn
\section{Quantum hardware experiment}
\label{ref:hardware}
\begin{figure}[!htb]
    \centering
    \includegraphics[width=0.95\linewidth]{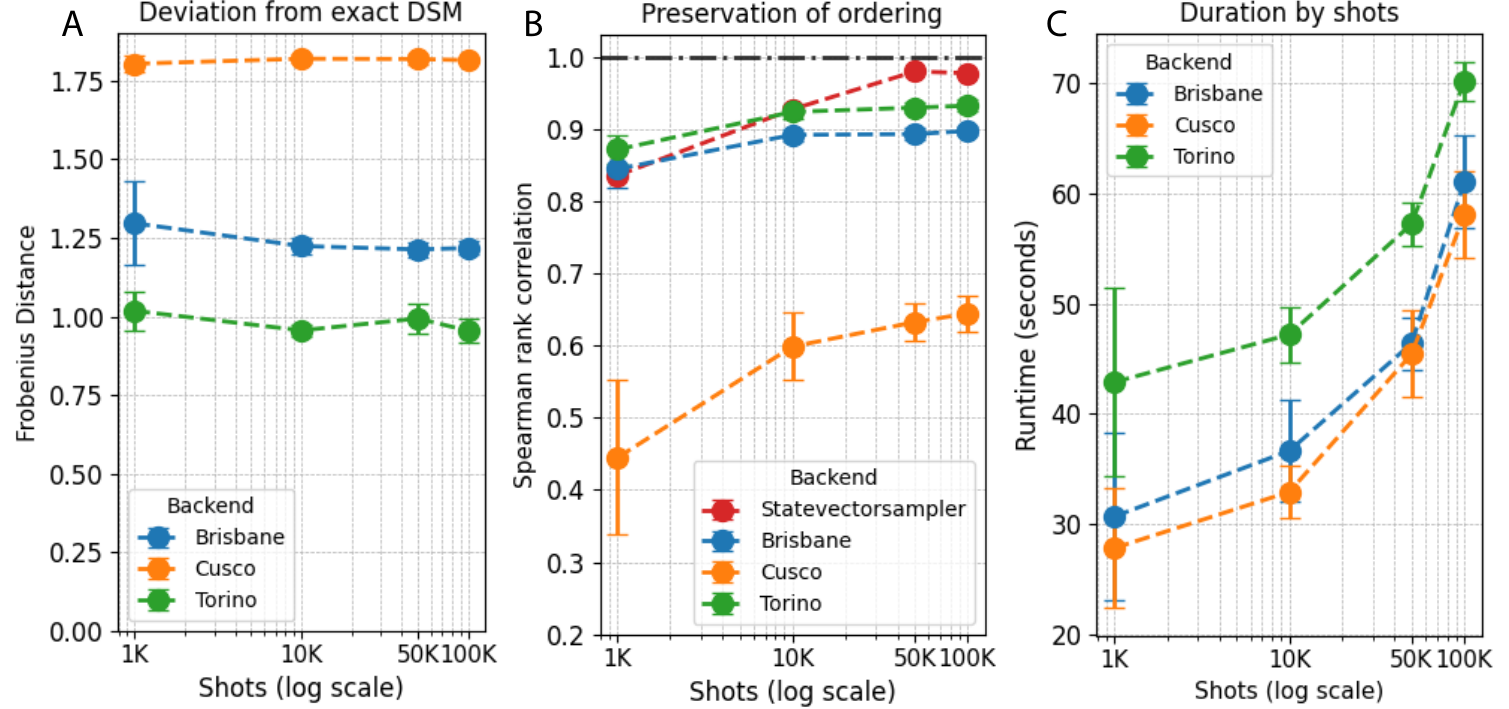}
    \caption{
    \textbf{Hardware experiment on different quantum computers available via \texttt{IBM Quantum Platform}.}
    \textbf{A} The Frobenius distance between the hardware-obtained DSM to its noise-free equivalent.
    % \textbf{B}: The entropy of the DSMs obtained from quantum hardware is significantly above the entropy of the exact DSM.
    \textbf{B}: The spearman rank correlation between the $64$ values in the noise-free and hardware-obtained DSMs show that the ordering of values is largely preserved.
    Statevectorsampler here denotes finite sampling from an ideal, noise-free statevector.
    %\textbf{D}: Execution time in seconds on different quantum computers.
    }
    \label{fig:hardware}
    \vspace{-2mm}
\end{figure}

We measured the extent and the effect of quantum hardware noise on a DSM produced by our quantum circuit. 
To that end, we picked a shallow circuit ($1$ layer) with $14$ qubits and computed the ground-truth $8\times 8$ DSM for a random input matrix through statevector simulation.
We then transpiled the circuit on three different quantum computers (Cusco, Brisbane and Torino, for details see~\autoref{tab:backends}) available to the public via the~\texttt{IBM Quantum Platform}.
\begin{table}[h!]
\centering
\caption{Comparison of IBM quantum backends (EPLG = Error Per Layer Gate).}
\begin{tabular}{l l c c}
\hline
\textbf{System} & \textbf{Architecture} & \textbf{Qubits} & \textbf{EPLG (\%)} \\
\hline
Torino   & Heron R1   & 133 & 1.3 \\
Brisbane & Eagle R3   & 127 & 2.2 \\
Cusco    & Eagle R3   & 127 & 6.8 \\
\hline
\end{tabular}
\label{tab:backends}
\end{table}
%
%
% \begin{figure*}[!htb]
%     \centering
%     \includegraphics[width=\linewidth]{figures/hardware_1l.pdf}
%     \caption{
%     \textbf{Hardware experiment on different quantum computers available via \texttt{IBM Quantum Platform}.}
%     \textbf{A} The Frobenius distance between the hardware-obtained DSM to its noisy-free equivalent.
%     \textbf{B}: The entropy of the DSMs obtained from quantum hardware is significantly above the entropy of the exact DSM.
%     \textbf{C}: The spearman rank correlation between the $64$ values in the noise-free and hardware-obtained DSMs show that the ordering of values is largely preserved.
%     \textbf{D}: Execution time in seconds as a function of shot count for different quantum computers.
%     }
%     \label{fig:hardware}
%     \vspace{-2mm}
% \end{figure*}
%
%
After using transpilation optimization level $1$, we obtained a circuit with a 2-qubit-depth of $15$ and and a total of $52$ two-qubit gates.
As error mitigation techniques, we used dynamical decoupling~\cite{ezzell2023dynamical}, Pauli twirling~\cite{wallman2016noise} and a projection to the Birkhoff polytope of the approximate-DSM obtained from the quantum circuit (see Section~\ref{sec:birkhoff}).
The results, shown in~\autoref{fig:hardware}, indicate that, consistently, \texttt{Cusco} was the noisiest machine and \texttt{Torino} yielded the best results. 
Moreover, in general, beyond a shot count of $10,000$ little performance improvement can be observed. 
This is a positive finding because it is substantially below the theoretical minimum given by the shot noise limit ($640,000$).
However, the deviation from the exact DSM, measured in Frobenius Distance, was substantial (\autoref{fig:hardware}\textbf{A}).
We analyzed the root cause of this and found that the deviation can be largely attributed to an increase in entropy.
%(\autoref{fig:hardware}\textbf{B}).
DSMs obtained from noisy quantum hardware converge toward the center of the Birkhoff polytope ($1/n$ in every cell).
The relative ordering of the absolute values instead is largely preserved (\autoref{fig:hardware}\textbf{C}).

\newpage

\section{Circuit execution times}
\begin{figure*}[!htbt]
    \centering
\includegraphics[width=0.8\linewidth]{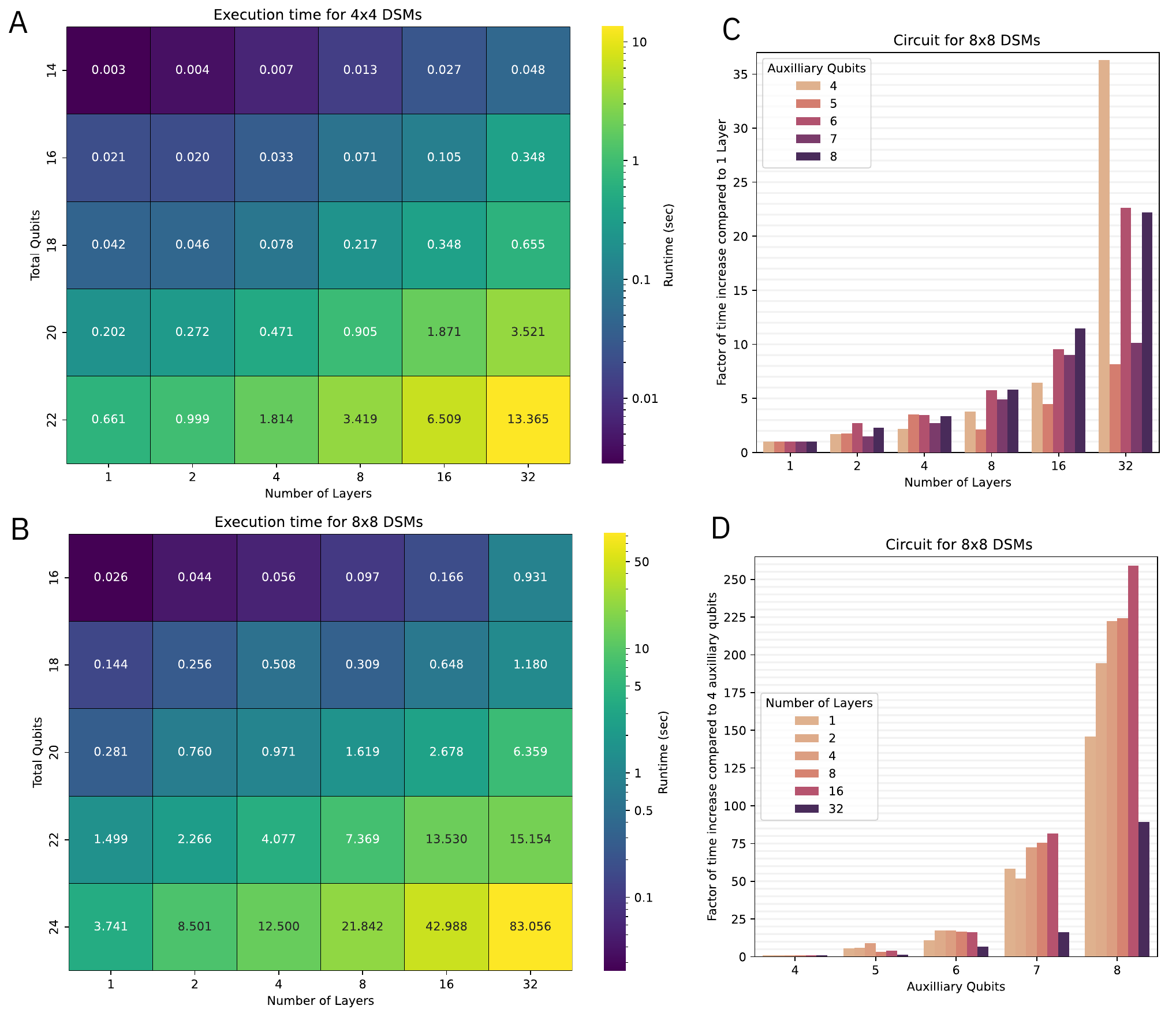}
    \caption{
    QontOT circuit execution times for DSM of size $4$ (\textbf{A}) and $8$ (\textbf{B}) for different combinations of qubits and circuit layers. 
    \textbf{C} and \textbf{D} show the relative increase in execution time as a function of increasing the number of layers (\textbf{C}), and qubits (\textbf{D}).
    Adding more layers has a sublinear effect on runtime, adding qubits requires exponential more runtime. 
    The minimal number of auxiliary qubits is $\log_2(n)+1$ and the total number of qubits is $2(q_d + q_a)$ where $q_d$ and $q_a$ are data and auxilliary qubits respectively.
    }
    \label{fig:runtime}
    \vspace{-2mm}
\end{figure*}

In~\autoref{fig:runtime} we report detailed runtimes for the QontOT algorithm for different combinations of circuit layers and auxiliary qubits.
\FloatBarrier
%
%
%
% \twocolumn
%
%

\FloatBarrier
\newpage
\clearpage
\section{Empirical circuit expressivity}

\begin{figure}[h]
    \centering
        \includegraphics[width=\columnwidth]{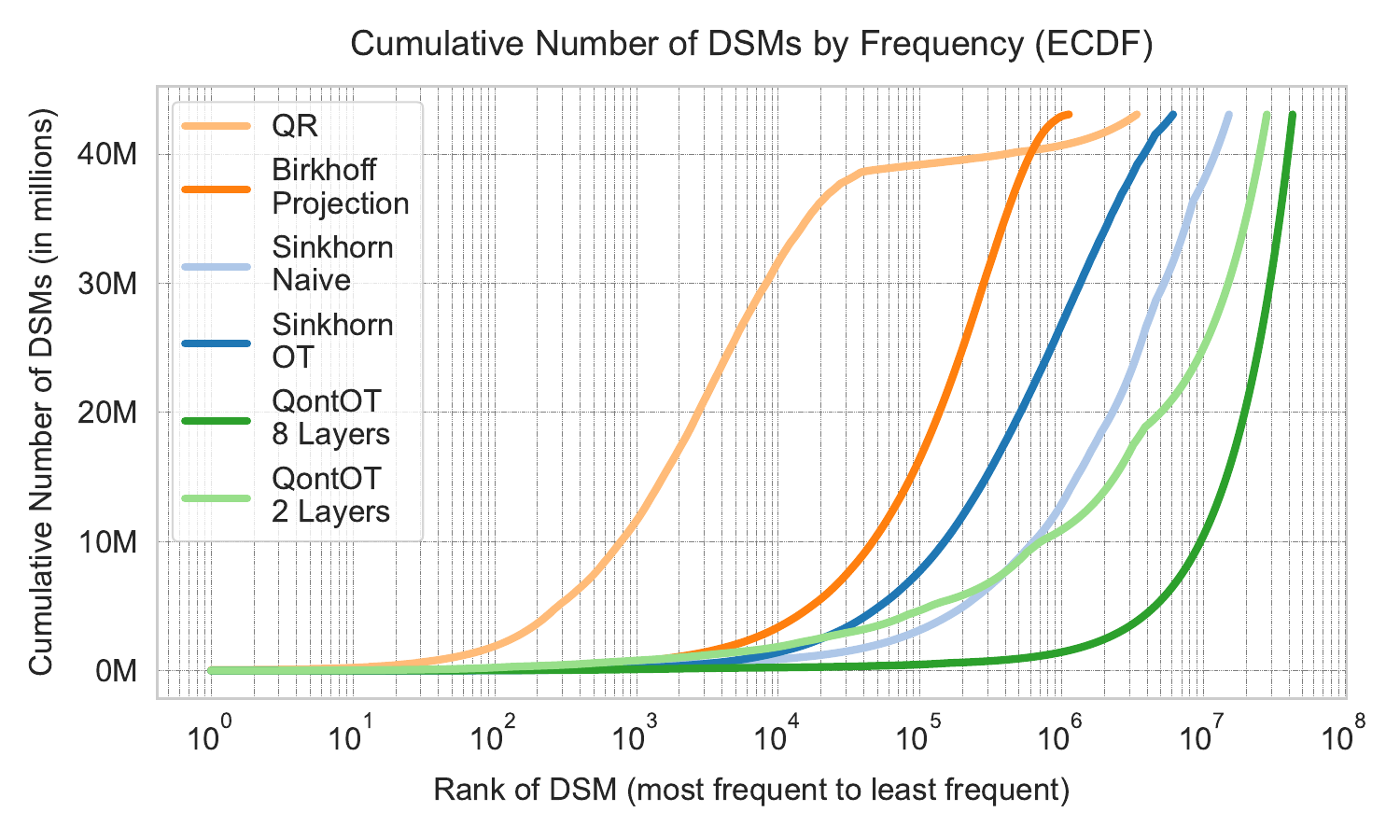}
        \caption{
 DSM counts were ranked descendingly and plotted against their cumulative count. 
    QontOT generally produces more diverse DSMs than Sinkhorn's algorithm.    
    }
        \label{fig:ecdf}
\end{figure}

\begin{figure}[h]
    \centering
        \includegraphics[width=0.7\columnwidth]{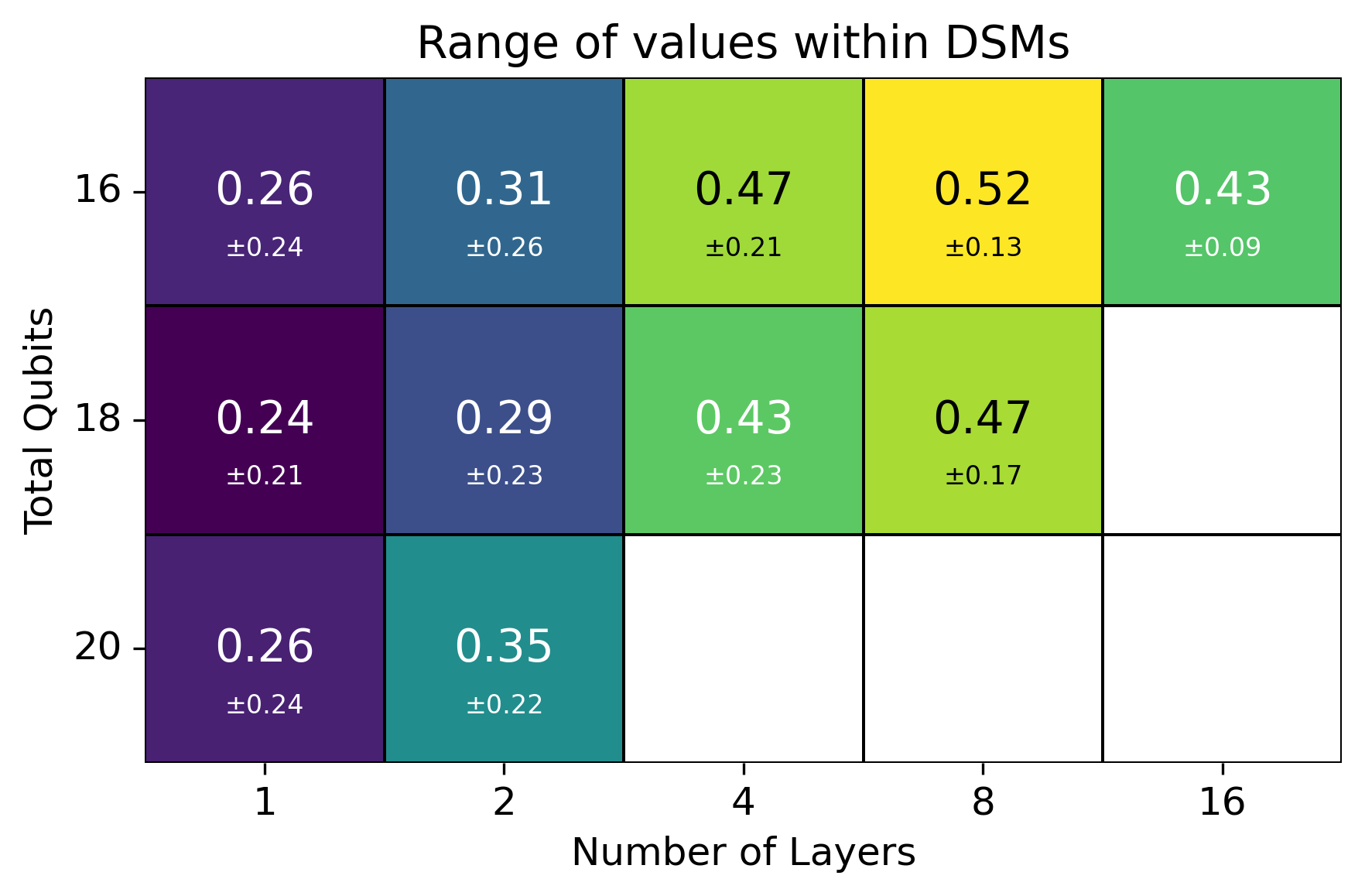}
        \caption{
        Mean range of observed values in the DSM obtained from a single, random input matrix, when randomly sampling $1000$ circuit parametrizations.
        }
        \label{fig:dsms_range}
\end{figure}

\FloatBarrier
\newpage
\section{QDSFormer results}
\subsection{Hyperparameters}
\label{sec:hyper}
For experiments on MNIST, FashionMNIST and the seven MedMNIST datasets, the ViT was configured with a hidden dimension of 128 and an MLP dimension expansion factor of 1. The model was tested with 1 to 4 Transformer layers, each containing a single attention head. No dropout was applied, and the batch size was set to 100. For the optimizer, Adam~\cite{Kingma2014AdamAM} was used, and the learning rate schedule followed the setup in \citet{sander2022sinkformers}, with an initial learning rate of 5e-4, decreasing by a factor of 10 at epochs 31 and 45.
For the more complex Eureka dataset, comprised of 56x56 RGB images, the hidden dimension was increased to 256, and a larger batch size of 512 was used. The MLP expansion factor was also doubled to 2. A cosine learning rate schedule was used with the optimizer AdamW~\cite{loshchilov2018adamW}. The scheduler uses 5 warmup epochs with a warm-up learning rate of 1e-6, the decay rate is set to 0.1 and the minimum learning rate is 1e-5, the other parameters follows the default TIMM settings \cite{rw2019timm}. For the optimizer the weight decay is 0.05 and betas $(0.9, 0.999)$.
All studied imaging datasets (MNIST, FashionMNIST and seven types of MedMNIST datasets) come with predefined train/validation/test splits. 
On the infrared spectral data of molecules from~\citet{alberts2025unraveling} we performed a 5-fold cross validation with $80\%/20\%$ train/test split. Hyperparameters were kept identical to previous experiments.
For the Eureka dataset, no Exponential Moving Average (EMA) is used.
Experiments were conducted on \texttt{POWER8} infrastructure in \texttt{Python 3.9} with \texttt{PyTorch}~\cite{paszke2017automatic} $1.13.1$ on machines with $16$ cores of \texttt{32GiB RDIMM DDR4 2.7 GHz}.
Due to the small size of the ViTs, training took between few hours and a day (for the slowest, i.e., end-to-end-differentiable configuration of the QDSFormer).
The Sinkformer~\cite{sander2022sinkformers} and the standard ViT implementation are taken from the original author's repository:~\url{https://github.com/michaelsdr/sinkformers}.
The results on the compositional Eureka dataset~\cite{hoffmann2024eureka} were generated with the ViT implementation of the original authors:~\url{https://github.com/boschresearch/eurekaMoments}.
The implementation of the QontOT circuit was implemented as described in~\citet{mariella2024quantum} and, as described in the main text, adapted to digest matrix (or vector) inputs rather than scalars only.

\subsection{QontOT ansatz types}
\begin{table}[!htb]
\centering
\caption{
Ablation study for a $2$-layer QDSFormer with different circuit ansatz types and varying number of layers on FashionMNIST. Mean/std of $5$ runs.
}
\tabcolsep=0.1cm
\small
\begin{tabular}{c|cccc}
\toprule
\textit{Circuit L.} & \textbf{Simple} & \textbf{Parted} & \textbf{Centrosymmetric} & \textbf{Trotter} \\
\midrule
1 & $\mathbf{88.0}_{\pm 0.10}$ & $87.7_{\pm 0.22}$  & $86.4_{\pm 0.23}$ & $87.7_{\pm 0.08}$   \\
8 & $\mathbf{89.9}_{\pm 0.15}$ & $89.8_{\pm 0.15}$  & $89.4_{\pm 0.17}$ & $88.4_{\pm 0.20}$  \\
\bottomrule
\end{tabular}
\label{tab:ansatztype}
\end{table}

\noindent \textbf{Simple}: This ansatz is the most generic and resembles a checkerboard structure formed by 4-parameter unit-blocks acting on two qubits each~\cite{khatri2019quantum,madden2022best}.
If all parameters are zero, it falls back to the identity. 
This ansatz is convenient because it is shallow in simulation but whose depth may vary depending on qubit layout of the quantum hardware. \\
\textbf{Parted}: This ansatz partitions the \textit{Simple} ansatz into two parts: $U = U_1 \otimes U_2$, where $U_2$ operates normally, and $U_1$ is transposed and placed around the initial Bell state. 
% This design reduces the circuit depth nearly by half, making it more efficient for quantum hardware. 
This design reduces the original \textit{Simple} ansatz circuit depth nearly by half, which may be more efficient on certain quantum hardware. However depending on the qubit layout of the quantum hardware, it carries the potential of increased transpiled circuit depth, as the two-qubit gates may act on distant qubits necessitating additional swap gates upon transpilation, which we observed on IBM Eagle and Heron quantum processing units. 
Unless mentioned otherwise, we used this ansatz in all our experiments as it yields shallower circuits in simulation and the increased depth compared to the \textit{Simple} ansatz was negligible at tested system sizes. \\
% Therefore, unless mentioned otherwise, we use this ansatz in all our experiments. \\
% While the circuit height is doubled, the reduced depth helps mitigate hardware noise and enables better performance on devices with limited coherence times.
\textbf{Centrosymmetric}: This was the predominantly used ansatz by~\citet{mariella2024quantum}. It is less generic, biasing toward properties of centrosymmetric matrices.\\
\textbf{Trotter}: This ansatz implements a second-order Trotter decomposition~\cite{smith2019simulating}. Each circuit layer corresponds to a Trotter step.

\FloatBarrier

\subsection{Time series classification}

\begin{table}[!htb]
    \centering
    \caption{
    Micro-F1 on IR spectra dataset across $5$-fold cross-validation with a 1-layer ViT.
    QDSFormer uses 16 circuit layers for both DSM sizes.
    % Micro-F1 on IR spectra in $5$-fold CV with 1-layer ViT (16 circuit layers).
    }
    % \vspace{-3mm}
    \tabcolsep=0.14cm
    % \small
    % \resizebox{1.0\linewidth}{!}{
    \begin{tabular}{c|ccccc}
    \toprule
\textbf{DSM} & \textbf{Softmax} & \textbf{Softmax$_{\sigma^2}$} & 
\textbf{QR} & 
\textbf{QDSFormer} & 
\textbf{Sinkhorn} \\
    \midrule
    $8\times8$ & $81.60_{\pm 0.34}$ & $81.41_{\pm 0.23}$ & $\underline{81.68}_{\pm 0.18}$ & $\textbf{81.70}_{\pm 0.05}$ & $81.38_{\pm 0.22}$ \\
    $16\times16$& $\textbf{81.55}_{\pm 0.07}$ & $80.94_{\pm 0.13}$ & $\underline{81.48}_{\pm 0.10}$ & $81.06_{\pm 0.27}$& $80.98_{\pm 0.30}$ \\
    \bottomrule
    \end{tabular}
    % }
    \label{tab:spectra}
\end{table}

\FloatBarrier
\subsection{Ablation studies}
\begin{table}[!htb]
    \centering
    \caption{QDSFormer ablation varying the circuit layers. 
    Exact numbers corresponding to~\autoref{fig:qdsmformer_layer_ablation}.
    %As baselines a 1-layer ViT is trained on MNIST and a 2-layered ViT is trained on FashionMNIST.
    }
    \begin{tabular}{l cc}
        \toprule
        \multirow{2}{*}{\textbf{Configuration}}  &\multicolumn{2}{c}{\textbf{Validation Accuracy (\%)}} \\ 
        &MNIST&FashionMNIST \\
        \midrule
        QDSFormer-1L & $83.4_{\pm 0.73}$ & $87.7_{\pm 0.22}$ \\%$81.0_{\pm 0.60}$\\
        QDSFormer-2L & $85.7_{\pm 0.60}$ & $88.5_{\pm 0.27}$\\%$82.0_{\pm 0.22}$\\
        QDSFormer-4L & $87.7_{\pm 0.73}$ & $89.3_{\pm 0.18}$\\%$83.6_{\pm 0.19}$ \\
        QDSFormer-8L & $91.8_{\pm 0.57}$ & $89.8_{\pm 0.15}$\\%$84.6_{\pm 0.07}$\\ 
        QDSFormer-16L & $93.8_{\pm 0.10}$ & $\textbf{90.0}_{\pm 0.15}$\\
        QDSFormer-32L & $\textbf{94.2}_{\pm 0.30}$ & $\textbf{90.0}_{\pm 0.13}$  \\
        \midrule
        \underline{Baseline} \\
        ViT & $92.9_{\pm 3.76}$ & $88.9_{\pm 0.12}$ \\
        %ViT & $89.1_{\pm 12.5}$ & $86.5_{\pm 0.19}$ \\
        %Sinkhorn  & $94.3_{\pm 1.97}$ & $84.2_{\pm 3.64}$\\ 
        %add baseline softmax?
        \bottomrule
    \end{tabular}
    \label{tab:circuit_layer_ablation}
\end{table}

\begin{figure}[!htb]
    \centering
    \includegraphics[width=0.7\linewidth]{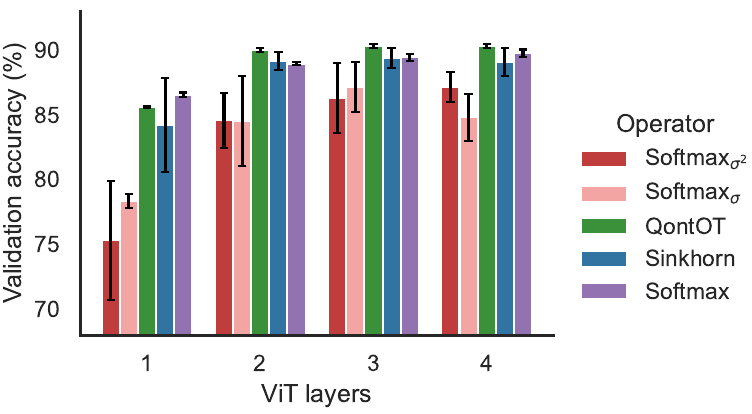}
    \caption{FashionMNIST results of different ViT layers for different attention types.
    }
    \label{fig:line_fashion}
    \vspace{-2mm}
\end{figure}

\begin{figure}[!htb]
    \centering
    \includegraphics[width=0.7\linewidth]{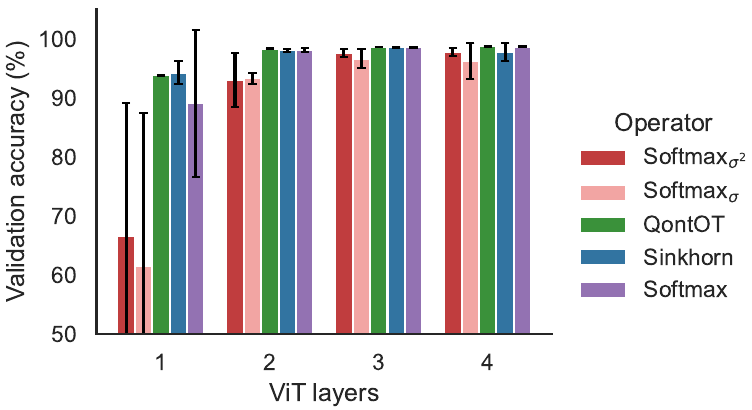}
    \caption{MNIST results of different ViT layers for different attention types.
    }
    \label{fig:line_mnist}
    \vspace{-2mm}
\end{figure}

\begin{comment}
\begin{table}[!htb]
    \centering
    \caption{
    Validation accuracy comparison between QR and QontOT doubly stochastic attention on FashionMNIST and MNIST.
    Mean and standard deviation are computed from $5$ training runs.
    }
    \label{tab:qr_ablation}
    \tabcolsep=0.10cm
    \small
    \begin{tabular}{r|cc|cc}
    \toprule
    Layers & \multicolumn{2}{c|}{\textbf{FashionMNIST}} & \multicolumn{2}{c}{\textbf{MNIST}} \\
    \cmidrule(lr){2-3} \cmidrule(lr){4-5}
    & \textbf{QR} & \textbf{QontOT} & \textbf{QR} & \textbf{QontOT} \\
    \midrule
    1 & \textbf{87.1}$_{\pm 0.26}$ & 84.8$_{\pm 0.17}$ & \textbf{96.6}$_{\pm 0.10}$ & 91.8$_{\pm 0.57}$ \\
    2 & 89.3$_{\pm 0.07}$ & \textbf{89.9}$_{\pm 0.15}$ & \textbf{98.3}$_{\pm 0.13}$ & \textbf{98.3}$_{\pm 0.08}$ \\
    3 & 89.4$_{\pm 0.11}$ & \textbf{90.2}$_{\pm 0.25}$ & \textbf{98.6}$_{\pm 0.13}$ & 98.1$_{\pm 0.06}$ \\
    4 & 89.5$_{\pm 0.07}$ & \textbf{90.3}$_{\pm 0.14}$ & \textbf{98.7}$_{\pm 0.11}$ & 98.4$_{\pm 0.06}$ \\
    \bottomrule
    \end{tabular}
\end{table}
\end{comment}

\begin{figure}[!htb]
    \centering
    \includegraphics[width=0.7\linewidth]{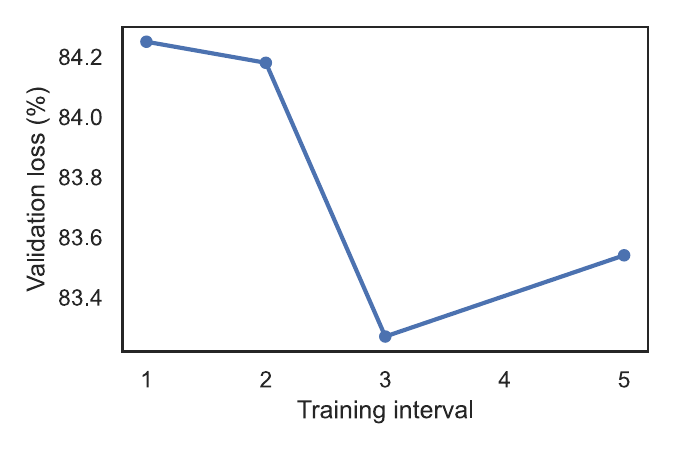}
    \caption{Ablation on different number of QontOT trainings on MNIST in the mixed circuit optimization strategy.
    }
    \label{fig:train_interval}
    \vspace{-2mm}
\end{figure}
\FloatBarrier

\subsection{Differentiable circuit}
\begin{figure}[!htb]
    \centering
    \begin{subfigure}[t]{0.49\linewidth}
        \centering
        \includegraphics[width=\linewidth]{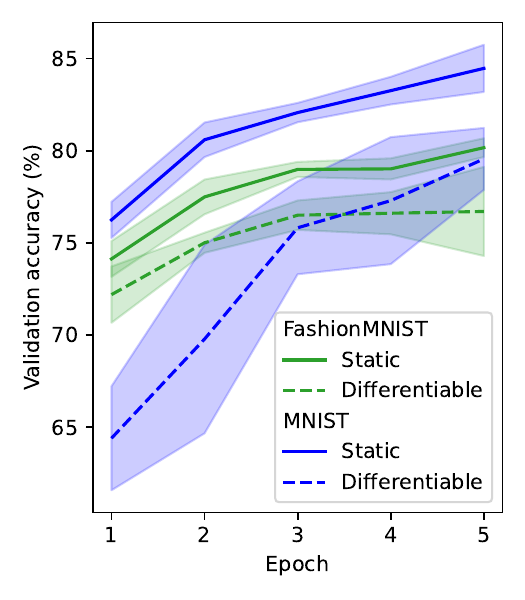}
        \caption{QDSFormer 8 circuit layers.
        }
        \label{fig:subfig_1layer}
    \end{subfigure}
    \hfill
    \begin{subfigure}[t]{0.49\linewidth}
        \centering
        \includegraphics[width=\linewidth]{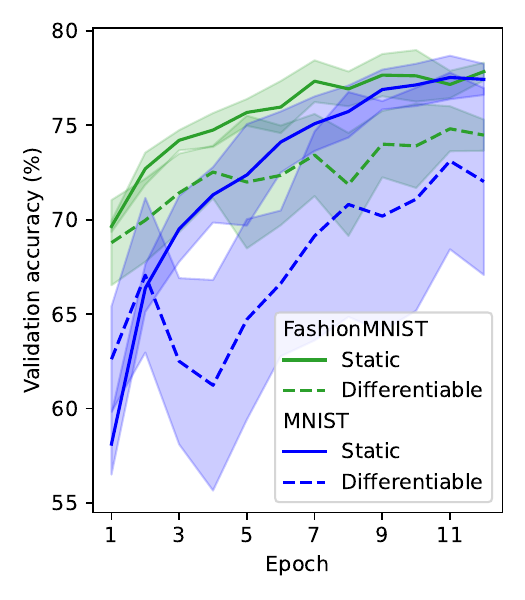}
        \caption{QDSFormer 1 circuit layers.
        }
        \label{fig:subfig_2layer}
    \end{subfigure}
    \caption{
    Differentiable QontOT training versus the static circuit. 
    }
    \label{fig:qdsformer_diff}
    \vspace{-2mm}
\end{figure}

\FloatBarrier

\newpage
\section{Counting DSMs}
\label{sec:dsm_calc}
In~\autoref{sec:full_3} we provide a full analytical solution to count the number of $3\times3$ DSMs for a given discretization $p\in \mathbb{N}$.
In~\autoref{sec:general_dsm} we strive to extend the analytic solution to an arbitrary $n \in \mathbb{N}$ but only provide a partial solution.
Finally, in~\autoref{sec:dsm_counts} we provide numerical results from brute-force calculation of the number of DSMs that verified the explicit analytical solution in~\autoref{sec:full_3}.

\subsection{Analytical solution for $n=3$}
\label{sec:full_3}
% In this work, we propose a method for calculating the number of DSMs for a given $n \times n$ matrix, where the matrix entries lie within a discretized range between 0 and 1. Our approach takes advantage of the fact that an $n \times n$ DSM is uniquely determined by its upper left $(n-1) \times (n-1)$ submatrix. The $(n-1) \times (n-1)$ matrices must satisfy two key constraints: (1) the row and column sums must be less than or equal to 1, and (2) the sum of all entries must be at least $n-2$. Our method first generates a discretized range between 0 and 1, then computes all possible combinations for the $(n-1) \times (n-1)$ submatrix. For each combination, we check if both constraints are satisfied. If the constraints are met, the values are used to construct the full $n \times n$ DSM. If not, we move to the next combination.

% We identified an analytical solution to calculate the total number of $3\times 3$ DSMs where every cell can take $ p $ values $\in [0,1]$.
% A $ 3 \times 3 $ matrix, it is uniquely determined by its upper left $ 2 \times 2 $ submatrix. 
\textit{Intuition.}
By systematically testing all combinations from the discretized range of values, starting with an initial $ 2 \times 2 $ zero matrix, each element is incrementally increased in the order $ x_{0,0}, x_{1,0}, x_{0,1}, x_{1,1} $ with the next highest value in the discretized range. Once an element reaches its maximum value, the next element is increased, and the preceding elements are reset to 0. This cycle repeats, starting again with the first element. 
%By examining the positions of submatrices that meet the constraints for forming DSMs, we identified the following relationship:

\textit{Explanation.}
Assume that \( n = 3 \) and a specific discretization \( p \in \mathbb{N}_+ \) are given.  
In this scenario, the corresponding \( 3 \times 3 \) matrix possesses $4$ degrees of freedom,  
implying that the associated submatrix has dimensions \( 2 \times 2 \). The first constraint requires that the sum of the elements in each row and each column of the matrix must not exceed 1.  

If a specific element \( e_{ij} \) with \( i, j \in \{0, 1\} \) is chosen and assigned a value \( x_i \),  
the possible values for the remaining elements in the same row and column can be determined.  

Given that each element can assume exactly \( p \) distinct values,  
the total number of combinations is computed as a sum over all \( p \) values:

\begin{equation}
f(3, p) = \sum_{i=1}^{p} c_i
\end{equation}

The possible values for the elements in the same row and column are restricted to the subset \( \{x_1, \dots, x_{p-i+1}\} \).  
As a result, the amount of submatrices that satisfy the first constraint can be expressed as:

\begin{equation}
f(3, p) = \sum_{i=1}^{p} \left[ \sum_{j=1}^{p-i+1} \left[ \sum_{k=1}^{p-i+1} c_{ijk} \right] \right]
\end{equation}

To determine the possible values for the last element,  
it is necessary to consider the elements \( e_{ij}' \).  
The minimum number of possible values derived from these elements defines the number of candidates for the last element:

\begin{equation}
f(3, p) = \sum_{i=1}^{p} \sum_{j=1}^{p-i+1}
\sum_{k=1}^{p-i+1}  
\sum_{l=1}^{\min(p-j+1, p-k+1)} \mathbbm{1}(i,j,k,l,p)
\end{equation}

Up to this point, only the first constraint has been considered.  
To fully satisfy the problem requirements, matrices that violate the second constraint must be excluded.  
The second constraint is satisfied when the sum of the indices of all elements does not exceed \( p \).  
Instead of subtracting \( 1 - \mathbb{I}(i, j, k, l, p) \), the condition is captured using an indicator function \( \mathbb{I}(i, j, k, l, p) \), defined as:

\begin{equation}
\mathbbm{1}(i, j, k, l, p) =  
\begin{cases}  
1 & \text{if } i + j + k + l - 3 \geq p, \\  
0 & \text{otherwise}.  
\end{cases}
\end{equation}

By incorporating \( \mathbbm{1}(i, j, k, l, p) \), the expression extends to:

\begin{equation}
f(3, p) = \sum_{i=1}^{p}  \sum_{j=1}^{p-i+1}  \sum_{k=1}^{p-i+1}  
 \sum_{l=1}^{\min(p-j+1, p-k+1)} \mathbbm{1}(i, j, k, l, p) 
\end{equation}

This can be summarized as follows:
\begin{equation}
f(3, p) = \sum_{(i, j, k, l) \in D(p)} \mathbbm{1}(i, j, k, l, p)
\end{equation}
where \(D(p) = \{(i, j, k, l) \mid 1 \leq i \leq p, 1 \leq j \leq p-i+1, 1 \leq k \leq p-i+1, 1 \leq l \leq \min(p-j+1, p-k+1)\}\).

This equation has been validated computationally for values up to \( p = 43 \),  
demonstrating alignment with our empirical results.

\subsection{General approximation}
\label{sec:general_dsm}
To determine the number of unique DSMs for a given $n,p\in \mathbb{N}$ we try to solve
\begin{equation}
f(n,p) = p^{(n-1)^2} - c_1 - c_2 + c_{12}
\end{equation}
where the first term calculates the number of DSM-candidate matrices, $c_1$ and $c_2$ measure how often the constraints are violated and $c_{12}$ is a small correction term counting cases where both constraints are violated.

Generally, $c_{12}$ is very small, yet difficult to compute, thus a tight lower bound can be given with the remaining three terms. 
Below, we provide a derivation for $c_2$. 
We leave the derivation of $c_1$ to future work.

\subsubsection{Constraint 2}
\setcounter{section}{3}
\setcounter{constraint}{1}
\begin{constraint}
\label{con:c2}
    The sum of the $n-1 \times n-1$ inner matrix must not be below $n-2$~\cite{chan1999volume}.
\end{constraint}
We aim to find a function $c_2(n,p)$ that computes the number of violations to~\autoref{con:c2} for a given $n,p \in \mathbb{N}$ when exhaustively looping over all $p^{(n-1)^2}$ candidate matrices that uniquely determine a $n \times n$ DSM. \\

An $n-1 \times n-1$ matrix where each cell $x_{ij}$ can take $p$ values has 
\begin{equation}
    |u| = (n-1)^2 (p-1) + 1
\end{equation}
unique possible sums.
These sums are regularly spaced from $0$ to $(n-1)^2$ with a step size of $p-1$, i.e.,
$u_i = \left\{ \frac{i}{p-1} \mid i \in \{0, 1, ..., |u|\}\right\}$.
This allows conversion to an integer problem (by multiplication of $p-1$) and apply \textit{Stars \& Bars Theorem 2}.

\setcounter{section}{4}
\begin{theorem}
\label{theo:sb2}
    For any $s,k\in \mathbb{N}$, the number of $k$-tuples $(x_0, ..., x_k)$ where $x_k\in \mathbb{N}_0$ with sum $s$ is equal to the number of multisets of cardinality $s$ taken from a set of size $k$:
    \begin{equation}
        \binom{s+k-1}{k-1}
    \end{equation}
\end{theorem}

Specifically, we set $k:=(n-1)^2$ and then define the set of sums that violate the constraint as 
$S := \{ s \in \mathbb{N}_0 \mid 0 \leq s \leq (n-2)(p-1) \}$.
Thus $|S| = (n-2)(p-1)$
We then compute the violations via:
% loop over all sums $s_i \in \{0, 1, (n-2)(p-1)\} \cap \mathbb{N}$ that violate the constraint:

\begin{equation}
    \label{eq:sb2_applied}
    \hat{c}_2(n,p) = \sum_{s=0}^{(n-2)(p-1)} \binom{s+(n-1)^2-1}{(n-1)^2-1}
\end{equation}

Unfortunately, this is only approximately correct because~\autoref{theo:sb2} assumes $x_k\in \mathbb{N}_0$, instead we require $x_i \in \{0, 1, \dots, p-1\}$.
Therefore, we exclude solutions where any $x_i > p-1$ through the inclusion exclusion principle as follows. \\

Assume that some $x_i>p-1$.
We set $\hat{x}_i=x_i - (p-1+1)$.
Since $\hat{x}_i\geq 0$, we can rewrite the original sum 
\begin{equation}
s = \sum x_i= \sum_{i\in I} \left( \hat{x_i} + p \right) + \sum_{i \notin I} x_i
\end{equation}
where $I$ is the set of indices where $x_i > p-1$.

Now let $m:=|I|$ be the number of violating variables, then
\begin{equation}
\label{eq:adj}
s -mp= \sum_{i\in I} \hat{x_i} + \sum_{i \notin I} x_i
\end{equation}

To find the number of non-negative integer solutions to~\autoref{eq:adj}, we can again leverage~\autoref{theo:sb2}, but now in a corrected form:

\begin{equation}
d(n,p,m,s) = 
\begin{cases} 
\binom{s - mp + (n - 1)^2 -1 }{(n - 1)^2 - 1} & \text{if } s - mp > 0, \\
0 & \text{otherwise}.
\end{cases}
\end{equation}

This gives a solution for a specific sum $s \in S$ and number of violations $m$. 
However, since $0\leq m \leq (n-1)^2$, we have to sum over all options of $m$ and apply the inclusion-exclusion principle to avoid over-/undercounting.

% Each candidate matrix must satisfy the two key constraints: (1) the row and column sums must not exceed 1, and (2) the sum of all entries must be at least $n-2$. Our method first generates a discretized range between 0 and 1, then computes all possible combinations for the $(n-1) \times (n-1)$ submatrix. 
% For each combination, we check if both constraints are satisfied. If the constraints are met, the values are used to construct the full $n \times n$ DSM. If not, we move to the next combination.
\begin{figure*}[!htb]
    \centering
        \includegraphics[width=\textwidth]{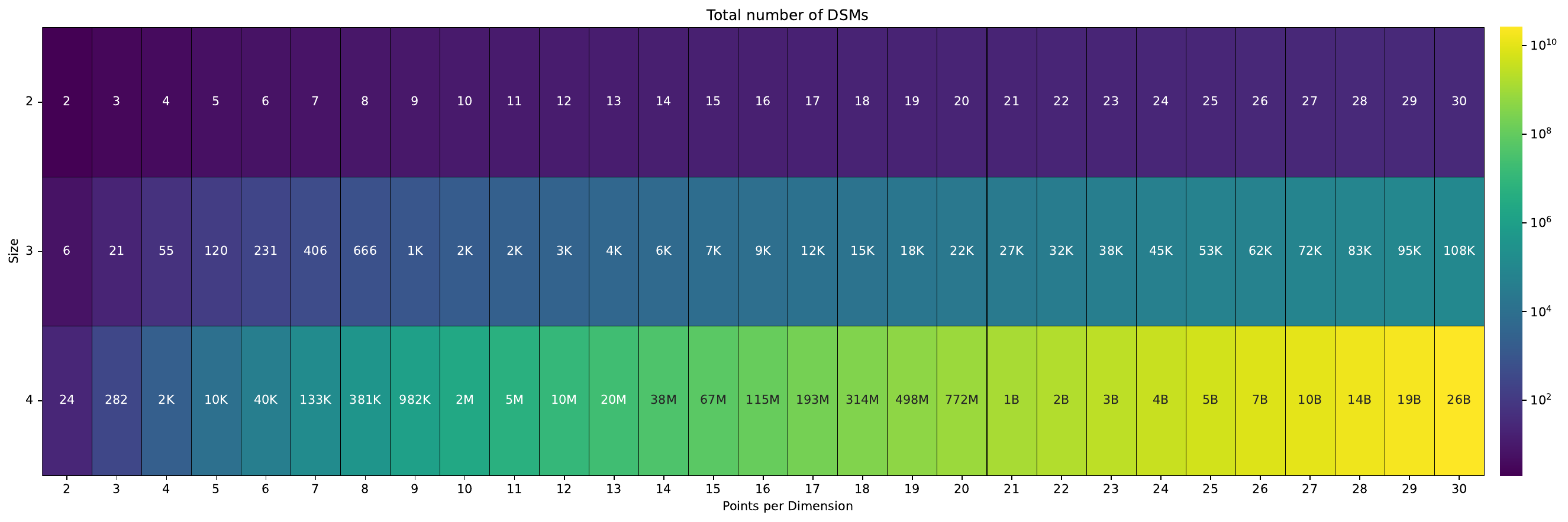}
        \caption{Number of DSMs of fixed size $n$ with a given number of discretization steps $p$; up to values of $n=4$ and $p=30$. }
        \label{fig:dsms_4d}
\end{figure*}

\begin{equation}
\label{eq:inclexcl}
c_2(n, p, s) = 
\sum_{m=0}^{(n-1)^2}
\left[
(-1)^m \binom{(n-1)^2}{m}
d(n,p,m,s)
\right]
\end{equation}
where $\binom{(n-1)^2}{m}$ accounts for the number of ways to choose $m$ out of $(n-1)^2$ variables that exceed $p-1$.

Plugging~\autoref{eq:inclexcl} back into the initial summation over all values $s \in S$ violating the constraint (see~\autoref{eq:sb2_applied}), we obtain the final formula:

\begin{multline}
        \label{eq:c2_final}
    c2(n,p) = \sum_{s=0}^{(n-2)(p-1)} c_2(n,p,s)  \\
    = \sum_{s=0}^{|S|} \sum_{m=0}^{(n-1)^2} (-1)^m \binom{(n-1)^2}{m}  
    \begin{cases} 
    0 & \hspace{-4em} \text{if } s - mp \leq 0,  \\
\binom{s - mp + n^2 - 2n}{n^2 - 2n} & \text{else}.
\end{cases}
\notag
\end{multline}
where $|S| = (n-2)(p-1)$.

\subsection{Empirical results}
\label{sec:dsm_counts}
To empirically determine the solutions to $f(n,p)$ we implemented a brute-force algorithm by iterating over all $p^{(n-1)^2}$ candidate matrices of size $(n-1) \times (n-1)$ and verifying whether the two constraints are not violated (see~\autoref{sec:dsm}).

\begin{figure}[!htb]
    \centering
        \includegraphics[width=0.7\columnwidth]{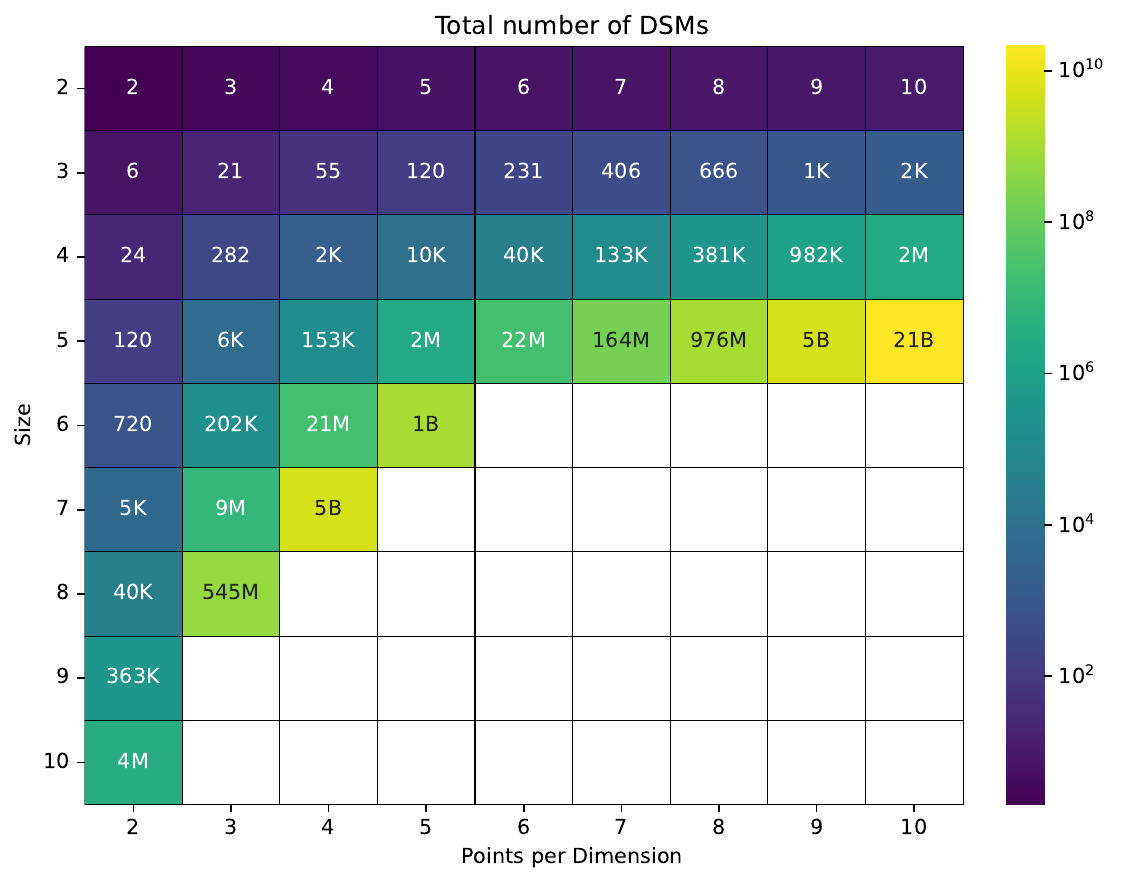}
        \caption{Number of DSMs of fixed size $n$ with a given number of discretization steps $p$; up to values of $n=10$ and $p=10$. 
        Empty cells require $>5$ days of compute time on a machine with $128$ cores and $128$GB RAM.
        }
        \label{fig:dsms_10d}
\end{figure}

The results are given in~\autoref{fig:dsms_10d} and~\autoref{fig:dsms_4d}.
Interestingly, $f(n,2)=n!$, but in general $f(n,p)$ scales super-factorially in $n$, for a given $p$.
\FloatBarrier
\section{Checklist information}

\begin{table}[ht]
\centering
\begin{tabular}{llll}
\toprule
\textbf{Dataset} & \textbf{Reference} & \textbf{License} & \textbf{Size} \\
\midrule
MNIST & \cite{lecun2010mnist} & GNU & 70,000 \\
Fashion-MNIST & \cite{xiao2017fashionmnistnovelimagedataset} & MIT & 70,000 \\
OCTMNIST & \cite{medmnistv1} & CC BY 4.0 & 109,000 \\
PneumoniaMNIST & \cite{medmnistv1} & CC BY 4.0 & 5,856 \\
TissueMNIST & \cite{medmnistv1} & CC BY 4.0 & 236,386 \\
OrganAMNIST & \cite{medmnistv1} & CC BY 4.0 & 58,830 \\
OrganCMNIST & \cite{medmnistv1} & CC BY 4.0 & 23,583 \\
OrganSMNIST & \cite{medmnistv1} & CC BY 4.0 & 25,211 \\
BreastMNIST & \cite{medmnistv1} & CC BY 4.0 & 780 \\
Compositional & \cite{hoffmann2024eureka} & GNU / MIT & 70,000 \\
IR Spectra & \cite{alberts2025unraveling} & CDLA & 790,000 \\
\bottomrule
\end{tabular}
\caption{Summary of datasets used, with references, licenses, and sizes.}
\end{table}

% \begin{table}[!htb]
% \centering
% \caption{Maximum validation accuracy after 100 epochs on the compositional dataset.}
% \begin{tabular}{cc}
% \toprule
% \textbf{Model} & \textbf{Accuracy ($\%$)} \\
% \midrule
% QDSFormer & $\textbf{70.1}_{\pm 14.00}$ \\
% QR & $\underline{61.3}_{\pm 11.10}$ \\
% ViT & $53.5_{\pm 0.09}$ \\
% Softmax$_{\sigma}$ & $49.3_{\pm 0.78}$ \\
% Sinkformer & $49.1_{\pm 0.85}$ \\
% Softmax$_{\sigma^2}$ & $48.6_{\pm 1.12}$ \\
% \bottomrule
% \end{tabular}
% \end{table}

\FloatBarrier
\newpage

\newpage
\clearpage
\section*{NeurIPS Paper Checklist}

\begin{enumerate}

\item {\bf Claims}
    \item[] Question: Do the main claims made in the abstract and introduction accurately reflect the paper's contributions and scope?
    \item[] Answer: \answerYes{} % Replace by \answerYes{}, \answerNo{}, or \answerNA{}.
    \item[] Justification: We report empirical evidence for all claims in the abstract and introduction.
    \item[] Guidelines:
    \begin{itemize}
        \item The answer NA means that the abstract and introduction do not include the claims made in the paper.
        \item The abstract and/or introduction should clearly state the claims made, including the contributions made in the paper and important assumptions and limitations. A No or NA answer to this question will not be perceived well by the reviewers. 
        \item The claims made should match theoretical and experimental results, and reflect how much the results can be expected to generalize to other settings. 
        \item It is fine to include aspirational goals as motivation as long as it is clear that these goals are not attained by the paper. 
    \end{itemize}

\item {\bf Limitations}
    \item[] Question: Does the paper discuss the limitations of the work performed by the authors?
    \item[] Answer: \answerYes{} % Replace by \answerYes{}, \answerNo{}, or \answerNA{}.
    \item[] Justification: The conclusion discusses limitations of our approach imposed by the noisy nature of current quantum hardware as well as the limited scaling to large-scale datasets.
    \item[] Guidelines:
    \begin{itemize}
        \item The answer NA means that the paper has no limitation while the answer No means that the paper has limitations, but those are not discussed in the paper. 
        \item The authors are encouraged to create a separate "Limitations" section in their paper.
        \item The paper should point out any strong assumptions and how robust the results are to violations of these assumptions (e.g., independence assumptions, noiseless settings, model well-specification, asymptotic approximations only holding locally). The authors should reflect on how these assumptions might be violated in practice and what the implications would be.
        \item The authors should reflect on the scope of the claims made, e.g., if the approach was only tested on a few datasets or with a few runs. In general, empirical results often depend on implicit assumptions, which should be articulated.
        \item The authors should reflect on the factors that influence the performance of the approach. For example, a facial recognition algorithm may perform poorly when image resolution is low or images are taken in low lighting. Or a speech-to-text system might not be used reliably to provide closed captions for online lectures because it fails to handle technical jargon.
        \item The authors should discuss the computational efficiency of the proposed algorithms and how they scale with dataset size.
        \item If applicable, the authors should discuss possible limitations of their approach to address problems of privacy and fairness.
        \item While the authors might fear that complete honesty about limitations might be used by reviewers as grounds for rejection, a worse outcome might be that reviewers discover limitations that aren't acknowledged in the paper. The authors should use their best judgment and recognize that individual actions in favor of transparency play an important role in developing norms that preserve the integrity of the community. Reviewers will be specifically instructed to not penalize honesty concerning limitations.
    \end{itemize}

\item {\bf Theory assumptions and proofs}
    \item[] Question: For each theoretical result, does the paper provide the full set of assumptions and a complete (and correct) proof?
    \item[] Answer: \answerYes{} % Replace by \answerYes{}, \answerNo{}, or \answerNA{}.
    \item[] Justification: In the appendix, we provide a full analytical solution to count the number of $3 \times 3$ DSMs for a given discretization. We then we strive to extend the analytic solution to an arbitrary $n \in \mathcal{N}$ but only find a partial solution.
    \item[] Guidelines:
    \begin{itemize}
        \item The answer NA means that the paper does not include theoretical results. 
        \item All the theorems, formulas, and proofs in the paper should be numbered and cross-referenced.
        \item All assumptions should be clearly stated or referenced in the statement of any theorems.
        \item The proofs can either appear in the main paper or the supplemental material, but if they appear in the supplemental material, the authors are encouraged to provide a short proof sketch to provide intuition. 
        \item Inversely, any informal proof provided in the core of the paper should be complemented by formal proofs provided in appendix or supplemental material.
        \item Theorems and Lemmas that the proof relies upon should be properly referenced. 
    \end{itemize}

    \item {\bf Experimental result reproducibility}
    \item[] Question: Does the paper fully disclose all the information needed to reproduce the main experimental results of the paper to the extent that it affects the main claims and/or conclusions of the paper (regardless of whether the code and data are provided or not)?
    \item[] Answer: \answerYes{} % Replace by \answerYes{}, \answerNo{}, or \answerNA{}.
    \item[] Justification: Experimental details are described in detail in the Hyperparameter subsection and the main body of the paper. 
    Our ViT implementations relied on previous, publicly available implementations (Sinkformer~\cite{sander2022sinkformers} and Eureka~\cite{hoffmann2024eureka}). 
    \item[] Guidelines:
    \begin{itemize}
        \item The answer NA means that the paper does not include experiments.
        \item If the paper includes experiments, a No answer to this question will not be perceived well by the reviewers: Making the paper reproducible is important, regardless of whether the code and data are provided or not.
        \item If the contribution is a dataset and/or model, the authors should describe the steps taken to make their results reproducible or verifiable. 
        \item Depending on the contribution, reproducibility can be accomplished in various ways. For example, if the contribution is a novel architecture, describing the architecture fully might suffice, or if the contribution is a specific model and empirical evaluation, it may be necessary to either make it possible for others to replicate the model with the same dataset, or provide access to the model. In general. releasing code and data is often one good way to accomplish this, but reproducibility can also be provided via detailed instructions for how to replicate the results, access to a hosted model (e.g., in the case of a large language model), releasing of a model checkpoint, or other means that are appropriate to the research performed.
        \item While NeurIPS does not require releasing code, the conference does require all submissions to provide some reasonable avenue for reproducibility, which may depend on the nature of the contribution. For example
        \begin{enumerate}
            \item If the contribution is primarily a new algorithm, the paper should make it clear how to reproduce that algorithm.
            \item If the contribution is primarily a new model architecture, the paper should describe the architecture clearly and fully.
            \item If the contribution is a new model (e.g., a large language model), then there should either be a way to access this model for reproducing the results or a way to reproduce the model (e.g., with an open-source dataset or instructions for how to construct the dataset).
            \item We recognize that reproducibility may be tricky in some cases, in which case authors are welcome to describe the particular way they provide for reproducibility. In the case of closed-source models, it may be that access to the model is limited in some way (e.g., to registered users), but it should be possible for other researchers to have some path to reproducing or verifying the results.
        \end{enumerate}
    \end{itemize}

\item {\bf Open access to data and code}
    \item[] Question: Does the paper provide open access to the data and code, with sufficient instructions to faithfully reproduce the main experimental results, as described in supplemental material?
    \item[] Answer: \answerNo{} % Replace by \answerYes{}, \answerNo{}, or \answerNA{}.
    \item[] Justification: 
    %At this point, public release of the code may enable competitors to produce similar results which may jeopardize our own publication goals in case this paper is rejected from NeurIPS. 
    While the entire development codebase for this project unfortunately cannot be made public at this point, specific parts of the code are available upon justified request.
    % We believe that reproducibility is key to make research impactful, thus we will make the code available upon acceptance of the paper.
    \item[] Guidelines:
    \begin{itemize}
        \item The answer NA means that paper does not include experiments requiring code.
        \item Please see the NeurIPS code and data submission guidelines (\url{https://nips.cc/public/guides/CodeSubmissionPolicy}) for more details.
        \item While we encourage the release of code and data, we understand that this might not be possible, so “No” is an acceptable answer. Papers cannot be rejected simply for not including code, unless this is central to the contribution (e.g., for a new open-source benchmark).
        \item The instructions should contain the exact command and environment needed to run to reproduce the results. See the NeurIPS code and data submission guidelines (\url{https://nips.cc/public/guides/CodeSubmissionPolicy}) for more details.
        \item The authors should provide instructions on data access and preparation, including how to access the raw data, preprocessed data, intermediate data, and generated data, etc.
        \item The authors should provide scripts to reproduce all experimental results for the new proposed method and baselines. If only a subset of experiments are reproducible, they should state which ones are omitted from the script and why.
        \item At submission time, to preserve anonymity, the authors should release anonymized versions (if applicable).
        \item Providing as much information as possible in supplemental material (appended to the paper) is recommended, but including URLs to data and code is permitted.
    \end{itemize}

\item {\bf Experimental setting/details}
    \item[] Question: Does the paper specify all the training and test details (e.g., data splits, hyperparameters, how they were chosen, type of optimizer, etc.) necessary to understand the results?
    \item[] Answer: \answerYes{} % Replace by \answerYes{}, \answerNo{}, or \answerNA{}.
    \item[] Justification: There is a dedicated section about hyperparameter choices in the appendix.
    \item[] Guidelines:
    \begin{itemize}
        \item The answer NA means that the paper does not include experiments.
        \item The experimental setting should be presented in the core of the paper to a level of detail that is necessary to appreciate the results and make sense of them.
        \item The full details can be provided either with the code, in appendix, or as supplemental material.
    \end{itemize}

\item {\bf Experiment statistical significance}
    \item[] Question: Does the paper report error bars suitably and correctly defined or other appropriate information about the statistical significance of the experiments?
    \item[] Answer: \answerYes{} % Replace by \answerYes{}, \answerNo{}, or \answerNA{}.
    \item[] Justification: All empirical experiments on the QDSFormer were repeatedly performed. 
    Error bars are shown in all plots and standard deviations are given in all tables.
    \item[] Guidelines:
    \begin{itemize}
        \item The answer NA means that the paper does not include experiments.
        \item The authors should answer "Yes" if the results are accompanied by error bars, confidence intervals, or statistical significance tests, at least for the experiments that support the main claims of the paper.
        \item The factors of variability that the error bars are capturing should be clearly stated (for example, train/test split, initialization, random drawing of some parameter, or overall run with given experimental conditions).
        \item The method for calculating the error bars should be explained (closed form formula, call to a library function, bootstrap, etc.)
        \item The assumptions made should be given (e.g., Normally distributed errors).
        \item It should be clear whether the error bar is the standard deviation or the standard error of the mean.
        \item It is OK to report 1-sigma error bars, but one should state it. The authors should preferably report a 2-sigma error bar than state that they have a 96\% CI, if the hypothesis of Normality of errors is not verified.
        \item For asymmetric distributions, the authors should be careful not to show in tables or figures symmetric error bars that would yield results that are out of range (e.g. negative error rates).
        \item If error bars are reported in tables or plots, The authors should explain in the text how they were calculated and reference the corresponding figures or tables in the text.
    \end{itemize}

\item {\bf Experiments compute resources}
    \item[] Question: For each experiment, does the paper provide sufficient information on the computer resources (type of compute workers, memory, time of execution) needed to reproduce the experiments?
    \item[] Answer: \answerYes{} % Replace by \answerYes{}, \answerNo{}, or \answerNA{}.
    \item[] Justification: Quantum circuit execution times are explicitly studied. Moreover we provide compute resource details in the hyperparameter section.
    \item[] Guidelines:
    \begin{itemize}
        \item The answer NA means that the paper does not include experiments.
        \item The paper should indicate the type of compute workers CPU or GPU, internal cluster, or cloud provider, including relevant memory and storage.
        \item The paper should provide the amount of compute required for each of the individual experimental runs as well as estimate the total compute. 
        \item The paper should disclose whether the full research project required more compute than the experiments reported in the paper (e.g., preliminary or failed experiments that didn't make it into the paper). 
    \end{itemize}
    
\item {\bf Code of ethics}
    \item[] Question: Does the research conducted in the paper conform, in every respect, with the NeurIPS Code of Ethics \url{https://neurips.cc/public/EthicsGuidelines}?
    \item[] Answer: \answerYes{} % Replace by \answerYes{}, \answerNo{}, or \answerNA{}.
    \item[] Justification: Code is respected.
    \item[] Guidelines:
    \begin{itemize}
        \item The answer NA means that the authors have not reviewed the NeurIPS Code of Ethics.
        \item If the authors answer No, they should explain the special circumstances that require a deviation from the Code of Ethics.
        \item The authors should make sure to preserve anonymity (e.g., if there is a special consideration due to laws or regulations in their jurisdiction).
    \end{itemize}

\item {\bf Broader impacts}
    \item[] Question: Does the paper discuss both potential positive societal impacts and negative societal impacts of the work performed?
    \item[] Answer: \answerNA{} % Replace by \answerYes{}, \answerNo{}, or \answerNA{}.
    \item[] Justification: This is a piece of foundational research in quantum machine learning that is currently only applicable to relatively small-scale data (i.e., small images). 
    Beyond the general societal implications of advances in quantum computing hardware, which will be vast and potentially disruptive, certainly for cryptography but potentially also for machine learning, we do not feel that there is anything specific about this paper.
    \item[] Guidelines:
    \begin{itemize}
        \item The answer NA means that there is no societal impact of the work performed.
        \item If the authors answer NA or No, they should explain why their work has no societal impact or why the paper does not address societal impact.
        \item Examples of negative societal impacts include potential malicious or unintended uses (e.g., disinformation, generating fake profiles, surveillance), fairness considerations (e.g., deployment of technologies that could make decisions that unfairly impact specific groups), privacy considerations, and security considerations.
        \item The conference expects that many papers will be foundational research and not tied to particular applications, let alone deployments. However, if there is a direct path to any negative applications, the authors should point it out. For example, it is legitimate to point out that an improvement in the quality of generative models could be used to generate deepfakes for disinformation. On the other hand, it is not needed to point out that a generic algorithm for optimizing neural networks could enable people to train models that generate Deepfakes faster.
        \item The authors should consider possible harms that could arise when the technology is being used as intended and functioning correctly, harms that could arise when the technology is being used as intended but gives incorrect results, and harms following from (intentional or unintentional) misuse of the technology.
        \item If there are negative societal impacts, the authors could also discuss possible mitigation strategies (e.g., gated release of models, providing defenses in addition to attacks, mechanisms for monitoring misuse, mechanisms to monitor how a system learns from feedback over time, improving the efficiency and accessibility of ML).
    \end{itemize}
    
\item {\bf Safeguards}
    \item[] Question: Does the paper describe safeguards that have been put in place for responsible release of data or models that have a high risk for misuse (e.g., pretrained language models, image generators, or scraped datasets)?
    \item[] Answer: \answerNA{} % Replace by \answerYes{}, \answerNo{}, or \answerNA{}.
    \item[] Justification: 
    \item[] Guidelines:
    \begin{itemize}
        \item The answer NA means that the paper poses no such risks.
        \item Released models that have a high risk for misuse or dual-use should be released with necessary safeguards to allow for controlled use of the model, for example by requiring that users adhere to usage guidelines or restrictions to access the model or implementing safety filters. 
        \item Datasets that have been scraped from the Internet could pose safety risks. The authors should describe how they avoided releasing unsafe images.
        \item We recognize that providing effective safeguards is challenging, and many papers do not require this, but we encourage authors to take this into account and make a best faith effort.
    \end{itemize}

\item {\bf Licenses for existing assets}
    \item[] Question: Are the creators or original owners of assets (e.g., code, data, models), used in the paper, properly credited and are the license and terms of use explicitly mentioned and properly respected?
    \item[] Answer: \answerYes{} % Replace by \answerYes{}, \answerNo{}, or \answerNA{}.
    \item[] Justification: A table with all datasets, citations, size and license terms are explicitly given in appendix.
    No data is re-distributed, license terms are respected.
    \item[] Guidelines:
    \begin{itemize}
        \item The answer NA means that the paper does not use existing assets.
        \item The authors should cite the original paper that produced the code package or dataset.
        \item The authors should state which version of the asset is used and, if possible, include a URL.
        \item The name of the license (e.g., CC-BY 4.0) should be included for each asset.
        \item For scraped data from a particular source (e.g., website), the copyright and terms of service of that source should be provided.
        \item If assets are released, the license, copyright information, and terms of use in the package should be provided. For popular datasets, \url{paperswithcode.com/datasets} has curated licenses for some datasets. Their licensing guide can help determine the license of a dataset.
        \item For existing datasets that are re-packaged, both the original license and the license of the derived asset (if it has changed) should be provided.
        \item If this information is not available online, the authors are encouraged to reach out to the asset's creators.
    \end{itemize}

\item {\bf New assets}
    \item[] Question: Are new assets introduced in the paper well documented and is the documentation provided alongside the assets?
    \item[] Answer: \answerNA{} % Replace by \answerYes{}, \answerNo{}, or \answerNA{}.
    % \item[] Justification: \justificationTODO{}
    \item[] Guidelines:
    \begin{itemize}
        \item The answer NA means that the paper does not release new assets.
        \item Researchers should communicate the details of the dataset/code/model as part of their submissions via structured templates. This includes details about training, license, limitations, etc. 
        \item The paper should discuss whether and how consent was obtained from people whose asset is used.
        \item At submission time, remember to anonymize your assets (if applicable). You can either create an anonymized URL or include an anonymized zip file.
    \end{itemize}

\item {\bf Crowdsourcing and research with human subjects}
    \item[] Question: For crowdsourcing experiments and research with human subjects, does the paper include the full text of instructions given to participants and screenshots, if applicable, as well as details about compensation (if any)? 
    \item[] Answer: \answerNA{} % Replace by \answerYes{}, \answerNo{}, or \answerNA{}.
    % \item[] Justification: \justificationTODO{}
    \item[] Guidelines:
    \begin{itemize}
        \item The answer NA means that the paper does not involve crowdsourcing nor research with human subjects.
        \item Including this information in the supplemental material is fine, but if the main contribution of the paper involves human subjects, then as much detail as possible should be included in the main paper. 
        \item According to the NeurIPS Code of Ethics, workers involved in data collection, curation, or other labor should be paid at least the minimum wage in the country of the data collector. 
    \end{itemize}

\item {\bf Institutional review board (IRB) approvals or equivalent for research with human subjects}
    \item[] Question: Does the paper describe potential risks incurred by study participants, whether such risks were disclosed to the subjects, and whether Institutional Review Board (IRB) approvals (or an equivalent approval/review based on the requirements of your country or institution) were obtained?
    \item[] Answer: \answerNA{} % Replace by \answerYes{}, \answerNo{}, or \answerNA{}.
    % \item[] Justification: \justificationTODO{}
    \item[] Guidelines:
    \begin{itemize}
        \item The answer NA means that the paper does not involve crowdsourcing nor research with human subjects.
        \item Depending on the country in which research is conducted, IRB approval (or equivalent) may be required for any human subjects research. If you obtained IRB approval, you should clearly state this in the paper. 
        \item We recognize that the procedures for this may vary significantly between institutions and locations, and we expect authors to adhere to the NeurIPS Code of Ethics and the guidelines for their institution. 
        \item For initial submissions, do not include any information that would break anonymity (if applicable), such as the institution conducting the review.
    \end{itemize}

\item {\bf Declaration of LLM usage}
    \item[] Question: Does the paper describe the usage of LLMs if it is an important, original, or non-standard component of the core methods in this research? Note that if the LLM is used only for writing, editing, or formatting purposes and does not impact the core methodology, scientific rigorousness, or originality of the research, declaration is not required.
    %this research? 
    \item[] Answer: \answerNA{} % Replace by \answerYes{}, \answerNo{}, or \answerNA{}.
    % \item[] Justification: \justificationTODO{}
    \item[] Guidelines:
    \begin{itemize}
        \item The answer NA means that the core method development in this research does not involve LLMs as any important, original, or non-standard components.
        \item Please refer to our LLM policy (\url{https://neurips.cc/Conferences/2025/LLM}) for what should or should not be described.
    \end{itemize}

\end{enumerate}

\end{document}